%% file: ms.tex
\documentclass[letterpaper, 10 pt, conference]{ieeeconf}  

\IEEEoverridecommandlockouts                              

\input{notation}

\renewcommand{\textfloatsep}{0em}

\usepackage{graphics} 
\usepackage{epsfig} 

\makeatletter
\let\NAT@parse\undefined
\makeatother

\usepackage{times} 
\usepackage{amsmath} 
\usepackage{amssymb}  

\usepackage[font=footnotesize]{caption}
\usepackage[font=footnotesize]{subcaption}
\usepackage{color}
\usepackage{hyperref}
\usepackage{algorithm}
\usepackage{algorithmic}
\usepackage[export]{adjustbox}

\def\secref#1{Sec.~\ref{#1}}

\def\figref#1{Fig.~\ref{#1}}

\def\eqref#1{Eq.~(\ref{#1})}

\def\image{{\mathbb{I}}}
\def\queryimage{{\image_q}}

\def\keypoint{{\bp}}

\def\descriptor{{\mathbf{d}}}
\def\querydescriptor{{\descriptor_q}}
\def\querydescriptors{{\left\{\descriptor_q\right\}}}

\def\inputdescriptor{{\descriptor_j}}
\def\inputdescriptors{{\left\{\inputdescriptor\right\}}}
\def\leaf{{\mathcal{L}}}
\def\leafs{{\left\{\leaf_i\right\}}}
\def\leafdescriptor{{\descriptor_{i,j}}}
\def\leafdescriptors{{\left\{\leafdescriptor\right\}}}
\def\descriptorsize{{\dim(\descriptor)}}
\def\images{{\left\{\image_i\right\}}}
\def\node{{\mathcal{N}}}
\def\nodes{{\left\{\node_i\right\}}}
\def\distanceimage{{e_\image}}
\def\distancedescriptor{{e_\descriptor}}
\def\dimdescriptor{{\dim(\descriptor)}}
\def\tree{{\mathcal{T}}}

\def\bitindex{{k_i}}
\def\bitindices{{\left\{k_i\right\}}}

\def\completeness{{c}}
\def\maxleafsize{{N_{max}}}

\input{gnuplot}

\title{\LARGE \bf HBST: A Hamming Distance embedding Binary Search Tree\\ for Visual Place Recognition}
\author{Dominik Schlegel and Giorgio Grisetti
\thanks{Both authors are with the Dept. of Computer, Control and Management Engineering,
        Sapienza University of Rome, Rome, Italy
        {\tt \{lastname\}@diag.uniroma1.it}}%
}

\begin{document}

\maketitle
\thispagestyle{empty}
\pagestyle{empty}

\begin{abstract}
  Reliable and efficient Visual Place Recognition is a major building block
  of modern SLAM systems. Leveraging on our prior work, in this paper
  we present a Hamming Distance embedding Binary Search Tree (HBST) approach
  for binary Descriptor Matching and Image Retrieval.
  HBST allows for descriptor Search and Insertion in logarithmic time
  by exploiting particular properties of binary Feature descriptors.
  We support the idea behind our search structure with a thorough
  analysis on the exploited descriptor properties and their
  effects on completeness and complexity of search and insertion.
  To validate our claims we conducted comparative experiments
  for HBST and several state-of-the-art methods
  on a broad range of publicly available datasets.
  HBST is available as a compact open-source C++ header-only library.
\end{abstract}

\section{Introduction}
\label{sec:introduction}
Visual Place Recognition (VPR) is a well known problem in Robotics and
Computer Vision~\cite{lowry-place-recognition-survey-2016} and
represents a building block of several applications in Robotics.
These range from Localization and Navigation to Simultaneous Localization and Mapping (SLAM).
The task of a VPR system is to localize an image within a database of places represented by other images.
VPR is commonly cast as a data association problem and used in loop closing modules of SLAM pipelines. 
A robust VPR system consists of one or multiple of the following components,
which progressively improve the solution accuracy:
\begin{itemize}
\item Image Retrieval: is the process of retrieving one or more \emph{images}
  from a database that are similar to a query one.
\item Descriptor Matching: consists of seeking \emph{points} between images which
  look similar. The local appearance of such points is captured by
  \emph{Feature descriptors}.
\item Geometric Verification: is a common pruning technique that removes points
  obtained from descriptor matching,
  which are inconsistent with the \emph{epipolar geometry}. 
\end{itemize}

In the domain of Image Retrieval,
common approaches compress entire images in single \emph{global} descriptors
to obtain high processing speed~\cite{2001-oliva-gist, dalal2005histograms}.
Recently, convolutional neural network methods demonstrated highly accurate results,
especially in the field of long-term VPR~\cite{2015-sunderhauf-convnet, arandjelovic2016netvlad, 2017arXiv170405016B}.
These methods, however, might suffer from a high ratio of false positives,
and thus often require a further stage of \emph{local} feature Descriptor Matching to reject wrong candidates.
Brute-force (BF) and k-d trees~\cite{kdtree} are two prominent methods for solving this task.

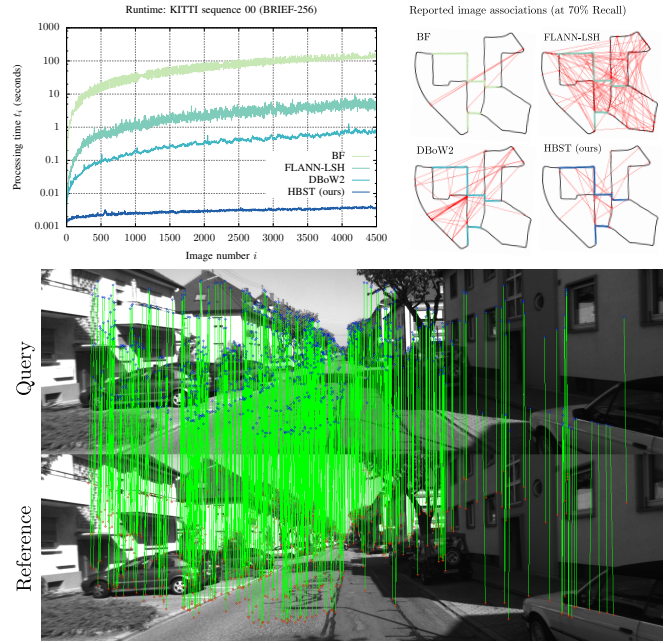
\begin{figure}
  \centering
  \vspace{0pt}
  \hspace{-5pt}
  \begin{subfigure}[t]{0.6\columnwidth}
    \resizebox{\columnwidth}{!}{\begin{picture}(7200.00,5040.00)%
    \gdef\gplbacktext{}
    \gdef\gplfronttext{}
    \gplgaddtomacro\gplbacktext{%
      \csname LTb\endcsname%
      \put(946,704){\makebox(0,0)[r]{\strut{}$0.001$}}%
      \csname LTb\endcsname%
      \put(946,1317){\makebox(0,0)[r]{\strut{}$0.01$}}%
      \csname LTb\endcsname%
      \put(946,1929){\makebox(0,0)[r]{\strut{}$0.1$}}%
      \csname LTb\endcsname%
      \put(946,2542){\makebox(0,0)[r]{\strut{}$1$}}%
      \csname LTb\endcsname%
      \put(946,3154){\makebox(0,0)[r]{\strut{}$10$}}%
      \csname LTb\endcsname%
      \put(946,3767){\makebox(0,0)[r]{\strut{}$100$}}%
      \csname LTb\endcsname%
      \put(946,4379){\makebox(0,0)[r]{\strut{}$1000$}}%
      \csname LTb\endcsname%
      \put(1078,484){\makebox(0,0){\strut{}$0$}}%
      \csname LTb\endcsname%
      \put(1714,484){\makebox(0,0){\strut{}$500$}}%
      \csname LTb\endcsname%
      \put(2350,484){\makebox(0,0){\strut{}$1000$}}%
      \csname LTb\endcsname%
      \put(2986,484){\makebox(0,0){\strut{}$1500$}}%
      \csname LTb\endcsname%
      \put(3622,484){\makebox(0,0){\strut{}$2000$}}%
      \csname LTb\endcsname%
      \put(4259,484){\makebox(0,0){\strut{}$2500$}}%
      \csname LTb\endcsname%
      \put(4895,484){\makebox(0,0){\strut{}$3000$}}%
      \csname LTb\endcsname%
      \put(5531,484){\makebox(0,0){\strut{}$3500$}}%
      \csname LTb\endcsname%
      \put(6167,484){\makebox(0,0){\strut{}$4000$}}%
      \csname LTb\endcsname%
      \put(6803,484){\makebox(0,0){\strut{}$4500$}}%
    }%
    \gplgaddtomacro\gplfronttext{%
      \csname LTb\endcsname%
      \put(176,2541){\rotatebox{-270}{\makebox(0,0){\strut{}Processing time $t_i$ (seconds)}}}%
      \put(3940,154){\makebox(0,0){\strut{}Image number $i$}}%
      \put(3940,4709){\makebox(0,0){\strut{}Runtime: KITTI sequence 00 (BRIEF-256)}}%
      \csname LTb\endcsname%
      \put(6219,2003){\makebox(0,0)[r]{\strut{}BF}}%
      \csname LTb\endcsname%
      \put(6219,1783){\makebox(0,0)[r]{\strut{}FLANN-LSH}}%
      \csname LTb\endcsname%
      \put(6219,1563){\makebox(0,0)[r]{\strut{}DBoW2}}%
      \csname LTb\endcsname%
      \put(6219,1343){\makebox(0,0)[r]{\strut{}HBST (ours)}}%
    }%
    \gplbacktext
    \put(0,0){\includegraphics{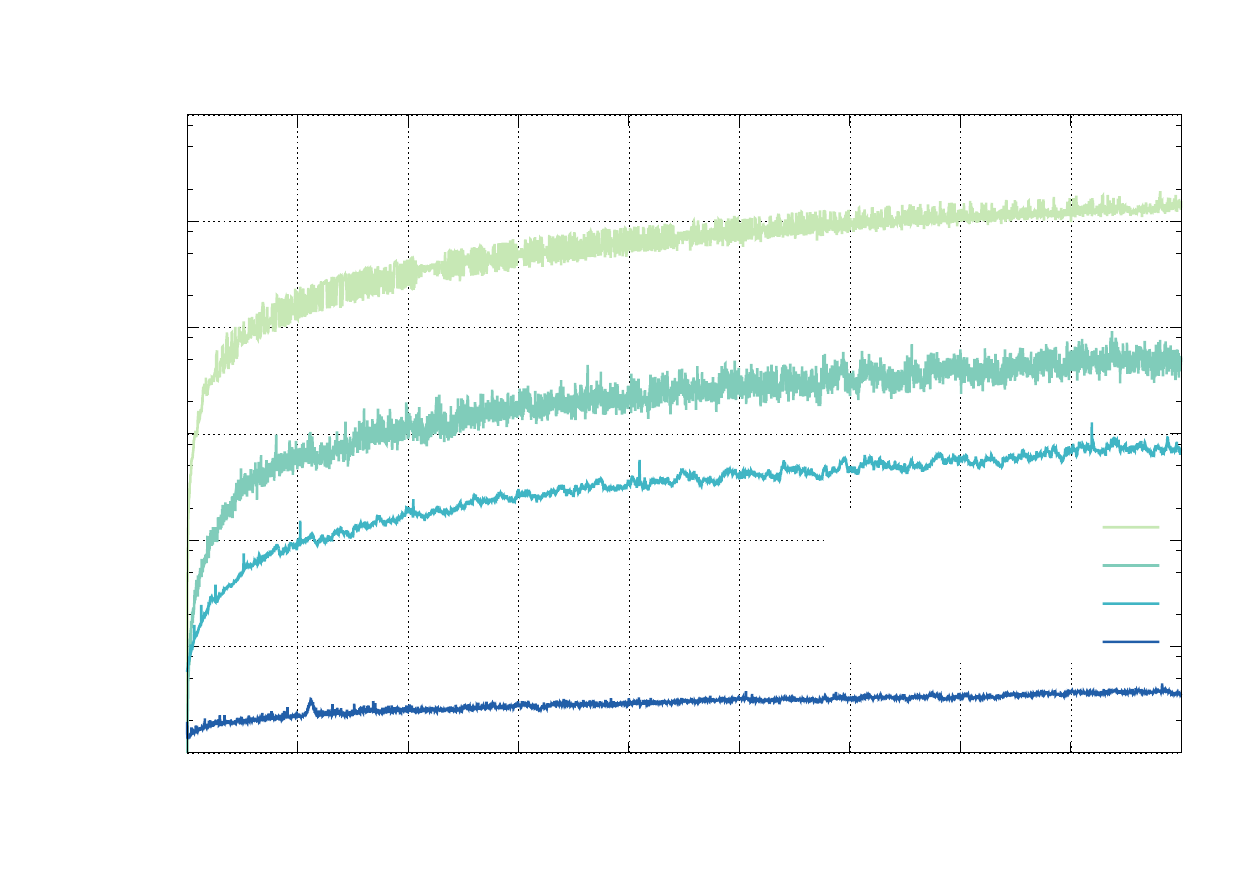}}%
    \gplfronttext
  \end{picture}}
    \vspace{-10pt}
  \end{subfigure}
  \hspace{-2pt}
  \begin{subfigure}[t]{0.39\columnwidth}
    \includegraphics[width=\columnwidth]{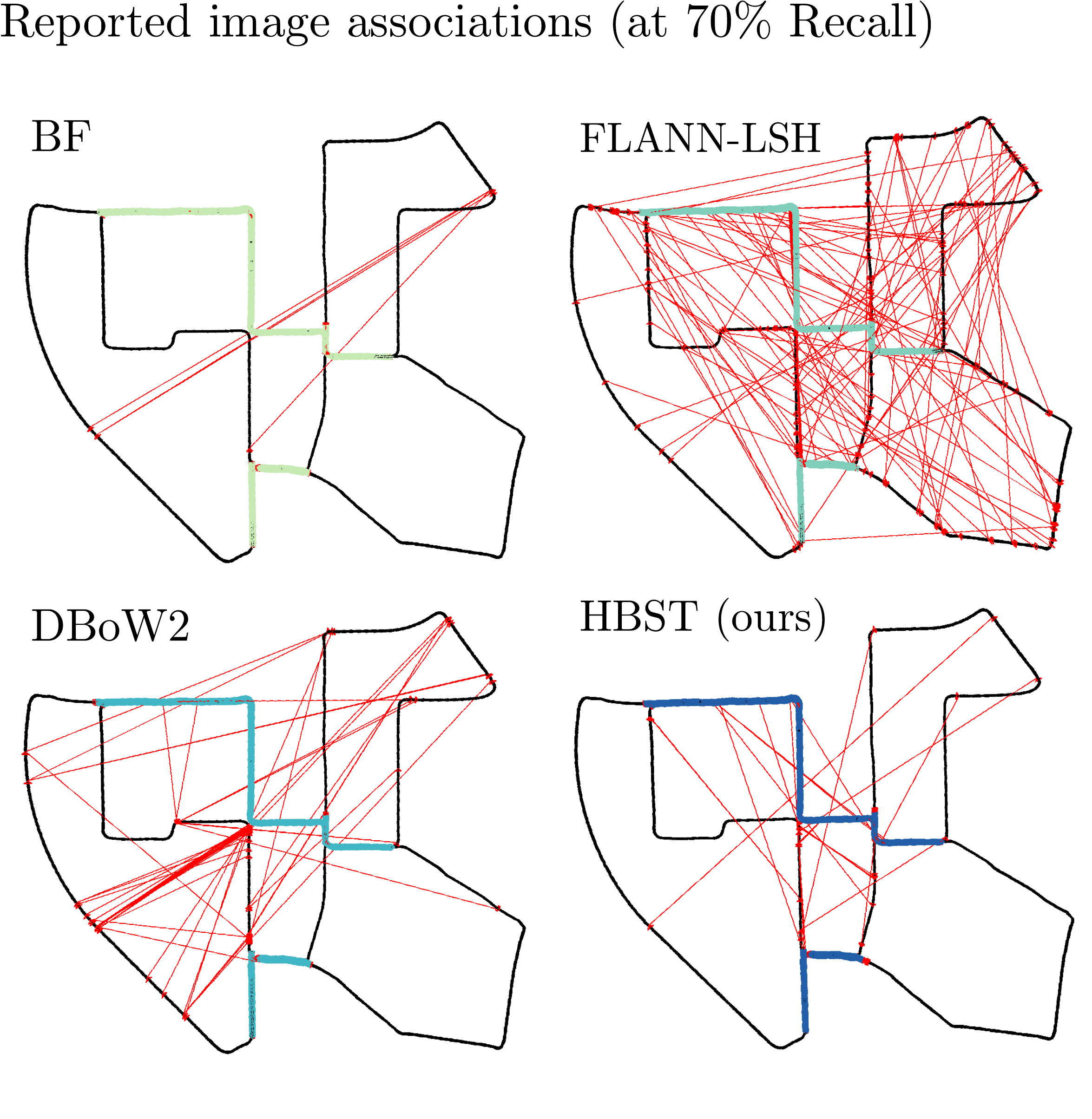}
    \vspace{-10pt}
  \end{subfigure}
  \begin{subfigure}{\columnwidth}
    \includegraphics[width=\columnwidth]{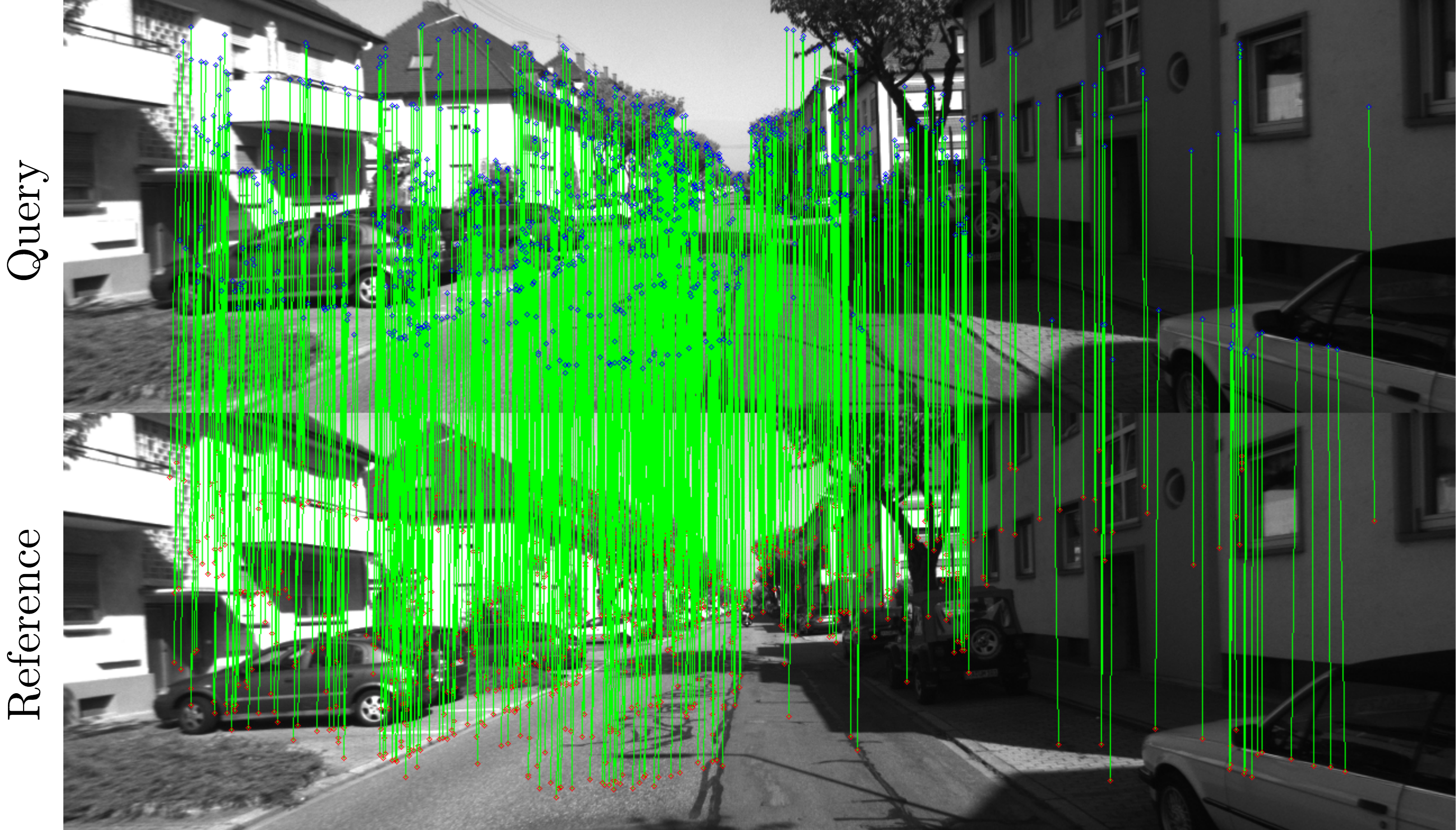}
  \end{subfigure}
  \vspace{-0pt}
  \caption{Matching performance of the proposed HBST approach on KITTI sequence 00.
  Top: Image processing times and image retrieval result of compared approaches at 70\% Recall.
  Bottom: A single query and reference image with highlighted descriptor matches provided by HBST.
  The shown query image was acquired 4'500 frames after the reference image.}
  \label{fig:motivation}
  \vspace{-0pt}
\end{figure}

Since the introduction of the BRIEF descriptor~\cite{calonder2010brief}, the computer vision community
embraced the use of binary descriptors due to their low computation and matching cost.
Many popular feature based SLAM systems such as ORB-SLAM~\cite{2017-mur-orbslam}
are built on these binary descriptors.

Whereas standard multi-dimensional search structures like k-d trees, are reported to perform well
for incremental construction with floating point descriptors like SURF~\cite{bay2008speeded},
the same approaches suffer a relevant performance loss when used with binary descriptors. 
This is the reason why in this work we focus on constructing a specific search structure,
that is tailored to matching \emph{binary} descriptors for VPR.

In this paper we propose an approach for binary descriptor matching
\emph{and} image retrieval that approximates the BF search.
Our system does not need to construct any kind of
dictionary and relies purely on a dynamically built binary search tree (BST)
that allows for logarithmic searches and insertions of binary descriptors.
Our approach runs several orders of magnitude faster than well-used implementations
of other state-of-the-art methods, while retaining high matching accuracy.
We provide our approach to the community in the form of
a compact C++ header-only library\footnote{{\scriptsize HBST is
available at: \url{www.gitlab.com/srrg-software/srrg_hbst}} \\
}
accompanied by several, straightforward use cases.

\section{Image Retrieval and Descriptor Matching}
In this section we discuss in detail the two fundamental building blocks of VPR
which we address in our approach:
Image Retrieval and Descriptor Matching.
We present related work directly in context of these two problems.

\subsection{Image Retrieval}
A system for \emph{image retrieval} returns the image $\image_i^\star$ contained in a database $\images$
that is the most similar to a given query image $\queryimage$ according to a similarity metric
$\distanceimage$. The more similar two images $\image_i$ and $\image_q$, the lower the resulting distance becomes.
More formally, image retrieval consists in solving the following problem:
\begin{equation}
  \image_i^\star = \argmin_{\image_i}\distanceimage(\image_q, \image_i) : \image_i \in \images.
  \label{eq:image-retrieval}
\end{equation}
Often one is interested in retrieving \emph{all} images in the database,
whose distance to the query image $\distanceimage$ is within a certain threshold $\tau_\image$:
\begin{equation}
  \left\{\image_i^\star\right\} = \left\{\image_i \in \images : \distanceimage(\image_q, \image_i) < \tau_\image \right\}.
  \label{eq:image-retrieval-output-multiple}
\end{equation}
The distance metric itself depends on the target application.
A straightforward example of distance between two images is the Frobenius norm
of the pixel-wise difference: 
\begin{equation}
  \distanceimage(\image_q, \image_i)=\left\|\queryimage-\image_i\right\|_F.
  \label{eq:distance-frobenius}
\end{equation}
This measure is not robust to viewpoint or illumination changes
and its computational cost is proportional to the image sizes.

\emph{Global image descriptors} address these issues by compressing an entire
image into a set of few values. In the remainder we will refer to a
global descriptor obtained from an image $\image$ as:
$\descriptor(\image)$. GIST of Olvia and Torralba~\cite{2001-oliva-gist} and Histogram of
Oriented Gradients (HOG) by Dalal and Triggs~\cite{dalal2005histograms} are two prominent
methods in this class.
GIST computes a whole image descriptor as the distribution
of different perceptual qualities and semantic classes detected in an
image. Conversely, HOG computes the descriptor as the histogram of
gradient orientations in portions of the image. 

When using global descriptors,
the distance between images is usually computed as the $L_2$ norm of
the difference between the corresponding descriptors:
\begin{equation}
  \distanceimage(\image_q, \image_i)=\left\|\descriptor(\image_q)-\descriptor(\image_i)\right\|_2.
  \label{eq:distance-global}
\end{equation}

Milford and Wyeth considered \emph{image sequences}
instead of single images for place recognition.
With SeqSLAM~\cite{milford2012seqslam} they presented an impressive SLAM system,
that computes and processes contrast enhancing image difference vectors between subsequent images.
Using this technique, SeqSLAM manages to recognize places that underwent heavy changes in appearance (e.g. from summer to winter).

In recent years, convolutional neural network approaches have shown to be very effective in VPR.
They are used to generate powerful descriptors that capture large portions of the scene at different resolutions.
For one, there is the CNN feature boosted SeqSLAM system of Bai~\emph{et al.}~\cite{2017arXiv170405016B},
accompanied by other off-the-shelf systems
such as ConvNet of S\"underhauf~\emph{et al.}~\cite{2015-sunderhauf-convnet}
or NetVLAD by Arandjelovi\'c~\emph{et al.}~\cite{arandjelovic2016netvlad}.
The large \emph{CNN descriptors} increase the description granularity and
therefore they are more robust to viewpoint changes than global descriptors.
CNN descriptors are additionally resistant to minor appearance changes,
making them suitable for \emph{lifelong} place recognition applications.
One can obtain up to a dozen CNN descriptors per image,
which enable for high-dimensional image distance metrics for $\distanceimage$.

However, if one wants to determine the \emph{relative location} at which images have been acquired, 
which is often the case for SLAM approaches,
additional effort needs to be spent. Furthermore, due to their holistic nature,
global descriptors might disregard the geometry of the scene and thus
are more likely to provide false positives.
Both of these issues can be handled by \emph{descriptor matching} and a subsequent geometric verification.

\subsection{Descriptor Matching}
Given two images $\queryimage$ and $\image_i$, we are interested in
determining which pixel $\keypoint_q \in \queryimage$ and which pixel
$\keypoint_j \in \image_i$, if any, capture the same point in the
world.  Knowing a set of these point correspondences, allows us to
determine the relative position of the two images up to a scale using
projective geometry~\cite{zisserman}. To this extent it is common to
detect a set of salient points $\{\keypoint\}$ (keypoints) in each
image.  Among others, the Harris corner
detector and the FAST detector are prominent approaches for
detecting keypoints. Keypoints are usually characterized by a strong
local intensity variation.

The local appearance around a keypoint $\keypoint$ is
captured by a descriptor $\descriptor(\keypoint)$
which is usually represented as a vector of either floating point or boolean values.
SURF~\cite{bay2008speeded} is a
typical floating point descriptor, while BRIEF~\cite{calonder2010brief}, BRISK~\cite{leutenegger2011brisk} and
ORB~\cite{rublee2011orb} are well known boolean descriptors.
The desired properties for local descriptors are the same as for global
descriptors: light and viewpoint invariance.
Descriptors are designed such that regions that appear locally similar in the
image result in similar descriptors, according to a certain metric
$\distancedescriptor$.
For floating point descriptors, $\distancedescriptor$ is usually chosen as the $L_2$-norm.
In the case of binary descriptors, the Hamming distance is a common choice.
The Hamming distance between two binary vectors is the number of bit changes needed to
turn one vector into the other, and can be effectively computed by current processors.

Finding the point $\keypoint_j^\star \in \image_i$  that is the most similar to a query $\keypoint_q \in \queryimage$
is resolved by seeking the descriptor $\descriptor(\keypoint^\star_j)$ with the minimum distance to the query $\descriptor(\keypoint_q)$:
\begin{equation}
 \keypoint_j^\star = \argmin_{\keypoint_j}\distancedescriptor(\descriptor(\keypoint_q), \descriptor(\keypoint_j)) : \keypoint_j\in \image_i.
 \label{eq:descriptor-matching}
\end{equation}
If a point $\keypoint_q \in \queryimage$ is not visible in $\image_i$,
\eqref{eq:descriptor-matching} will still return a point
$\keypoint_j^\star \in \image_i$.  Unfeasible matches however will
have a high distance, and can be rejected whenever their distance
$\distancedescriptor$ is greater than a certain \emph{matching threshold} $\tau$.
The most straightforward way to compute~\eqref{eq:descriptor-matching} is the brute-force (BF) search.
BF computes the distance $\distancedescriptor$ between $\keypoint_q$ and \emph{every} $\keypoint_j\in \image_i$.
And hence \emph{always} returns the \emph{closest} match for each query.
This unbeatable accuracy comes with a computational
cost proportional to the number of descriptors $N_\descriptor = \left|\{\descriptor(\keypoint_j)\}\right|$.
Assuming $N_\descriptor$ is the average number of
descriptors extracted for each image, finding the best correspondence
for each keypoint in the query image would require
$\complexity(N_\descriptor^2)$ operations. In current applications,
$N_\descriptor$ ranges from 100 to 10'000, hence using BF for descriptor
matching quickly becomes computationally prohibitive. 

To carry on the correspondence search in a more effective way it is common to organize
the descriptors in search structure, typically a tree. In the case of
floating point descriptors, FLANN (Fast Approximate Nearest Neighbor Search Library) of Muja and Lowe~\cite{flann}
with k-d tree indexing is a common choice.
When working with binary descriptors,
the (Multi-Probe) Locality-sensitive hashing (LSH)~\cite{lsh} by Lv~\emph{et al.}
and hierarchical clustering trees (HCT) of Muja and Lowe~\cite{2012-muja-fast-matching}
are popular methods to index the descriptors with FLANN.
While LSH allows for \emph{database incrementation} at a decent computational cost,
HCT quickly exceeds real-time constraints.
Accordingly, we consider only LSH in our result evaluations.
The increased speed of FLANN compared to BF comes at a decreased accuracy of the matches.
In our previous work~\cite{schlegel2016visual} we presented a binary search tree structure,
that resolves~\eqref{eq:descriptor-matching} in logarithmic time. However,
a tree had to be built for every desired image candidate pair.

\subsection{Image Retrieval based on Descriptor Matching}
\label{sec:image-retrieval-descriptor-matching}
Assuming to have an efficient method to perform descriptor matching as defined in~\eqref{eq:descriptor-matching},
one could design a simple, yet effective image retrieval system by
using a voting scheme.
An image $\image_i$ will receive at most one vote $\left<\keypoint_q, \keypoint^\star_{i,j}\right>$ for each keypoint $\keypoint_q \in \queryimage$
that is successfully matched with a keypoint of another image $\keypoint^\star_{i,j} \in \image_i$.
The distance between two images $\queryimage$ and $\image_i$ is then the number of votes
$\left|\left\{\left<\keypoint_q, \keypoint^\star_{i,j}\right>\right\}\right|$
normalized by the number of descriptors in the query image $N_\descriptor$:
\begin{equation}
  \distanceimage(\image_q, \image_i)=\frac{\left|\left\{\left<\keypoint_q, \keypoint^\star_{i,j}\right>\right\}\right|}{N_\descriptor}.
  \label{eq:voting-scheme}
\end{equation}
The above procedure allows to gather reasonably good matches at a cost
proportional to both, the number of descriptors in the query image $N_\descriptor$ and
the cost of retrieving the most likely descriptor as defined in~\eqref{eq:descriptor-matching}.

An alternative strategy to enhance the efficiency of image retrieval,
when local descriptors are available, is to compute \emph{a single}
image ``descriptor'' from \emph{multiple} feature descriptors. Bag-of-visual-Words (BoW)
approaches follow this strategy by computing an image descriptor as
the \emph{histogram of the distribution of words} appearing in the image. A
\emph{word} represents a group of nearby descriptors, and is learned
by a clustering algorithm such as k-means from a set of train
descriptors. 
To compute the
histogram, each keypoint is converted in a set of weights computed as
the distance of the descriptors' keypoint from the centroid of each
word in the dictionary. The histogram is then normalized by the sum
of word weights in the scene.
Images that present similar distribution of words are likely to be
similar.  Comparing a pair of images can be done in a time linear in
the number of words in the dictionary.  This procedure has been shown
to be both robust and efficient, however it does not provide
point correspondences, that are required for geometric verification.

Notably the open-source library DBoW2 by Galvez-Lopez and Tardos~\cite{galvez2012bags} extends the data
structures used in BoW to add point correspondences to the system.
This is done by storing an \emph{Inverted Index} from words to
descriptors that are close to a specific word.  To retrieve the
keypoints $\keypoint^\star_j$ that are similar to a query
$\keypoint^\star_j$ one can pick the words in the
dictionary that are best represented by
$\descriptor(\keypoint^\star_q)$ and from them retrieve the
descriptors through the inverted index.
In the current version of DBoW2,
Galvez-Lopez and Tardos provide also a \emph{Direct Index} descriptor index
for correspondence access.

DBoW2 is integrated within the recently published
ORB-SLAM2~\cite{2017-mur-orbslam} by Mur-Artal~\emph{et al.} and displays fast
and robust performance for ORB descriptors~\cite{rublee2011orb}.
Another famous BoW based approach is FAB-MAP~\cite{OpenFabMap}
developed by Cummins~\emph{et al.}
FAB-MAP allows to quickly retrieve similar images on very large
datasets. 
FAB-MAP uses costly SURF descriptors~\cite{bay2008speeded} to maintain a certain level of individuality
between the massive number of images described.
Typically, BoW is used to determine a preliminary set of good image candidates,
on which BF, FLANN or BST descriptor matching is performed.
This is a common practice for SLAM systems,
that require high numbers of matches for few images.

In this paper, we present a novel approach that:
\begin{itemize}
\item Allows to perform image retrieval and descriptor matching \emph{with} correspondences
  faster than BoW approaches perform image retrieval \emph{without} correspondences.
\item Yields levels of search correctness and completeness comparable to the one achieved
  by to state-of-the-art methods such as FLANN-LSH~\cite{lsh} and DBoW2~\cite{galvez2012bags}.
\item Allows for incremental insertion of subsequent descriptor sets (i.e. images)
  in a time bounded by the dimension of the descriptors $\descriptorsize$.
\end {itemize}
Furthermore, we provide our approach as a compact C++ header-only
library, that does not require a vocabulary or any other pretrained information.
The library is accompanied by a set of simple use cases
and includes an OpenCV wrapper.

\section{Our Approach}
We arrange binary feature descriptors $\inputdescriptors$ extracted
from each image $\image_i$ of an image sequence $\images$ in a binary tree.
This tree allows us to efficiently perform descriptor matching. 
Additionally, we build a voting scheme on top of this method that enables fast and robust image retrieval.

\subsection{Tree Construction}
\label{sec:approach-tree}
\begin{figure}[ht!]
  \centering
  \vspace{-5pt}
  \includegraphics[width=\columnwidth]{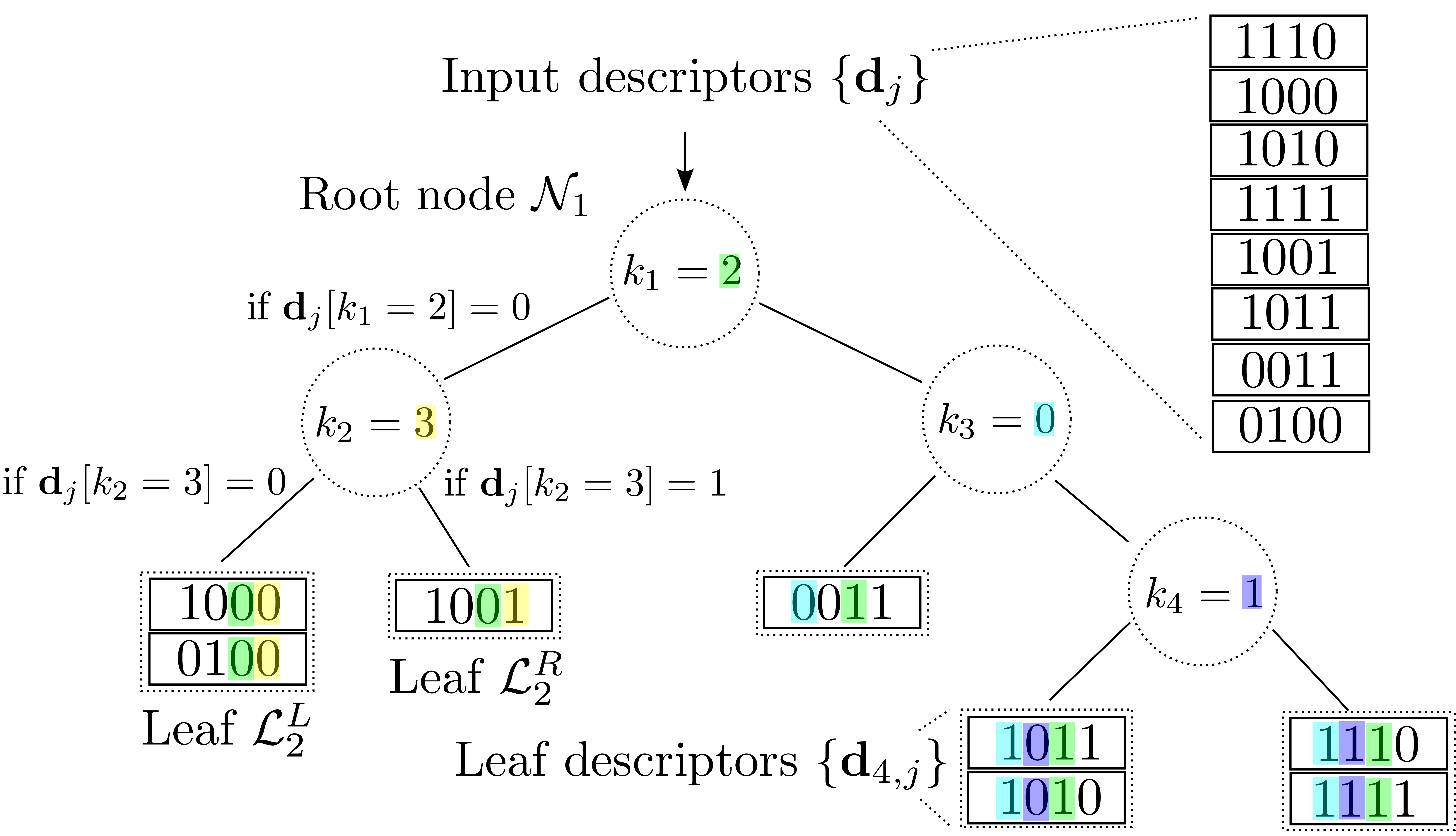}
  \vspace{-10pt}
  \caption{HBST tree construction for a scenario with 8 input descriptors of dimension $\descriptorsize = 4$.
  The tree contains 4 nodes $\nodes$ (circles) with bit indices $\left\{\bitindex\right\}$,
  5 leafs $\leafs$ (rectangles) and has maximum depth $h=3$.}
  \label{fig:hbst-construction}
  \vspace{-5pt}
\end{figure}

In our tree, each leaf $\leaf_i$ stores a subset $\leafdescriptors$ of the input descriptors $\inputdescriptors$.
The leafs partition the input set such that each descriptor $\inputdescriptor$ belongs to a single leaf.
Every non-leaf node $\node_i$ has exactly two children and stores an index $\bitindex \in [0, .., \descriptorsize-1]$.
Where $\descriptorsize$ is the descriptor dimension,
corresponding to the number of bits contained in each descriptor.
We require that in each path from the root to a leaf a specific index value $\bitindex$ should appear at most once.
This limits the depth of the tree $h$ to the dimension of the descriptors.
\figref{fig:hbst-construction}  illustrates an example tree constructed from 8 binary input descriptors $\inputdescriptors$ according to these rules.

A descriptor $\inputdescriptor$ is stored in the left or in the right subtree depending
on $\inputdescriptor[\bitindex]$, that is the bit value of
$\inputdescriptor$ at index $\bitindex$. The structure of the tree is
determined by the choice of the bit indices $\bitindices$ in the intermediate nodes
and by the respective number of descriptors stored in the leafs $\leafs$.

\subsection{Descriptor Search and Matching}
\label{sec:approach-search}
\begin{figure}[ht!]
  \centering
  \vspace{-0pt}
  \includegraphics[width=\columnwidth]{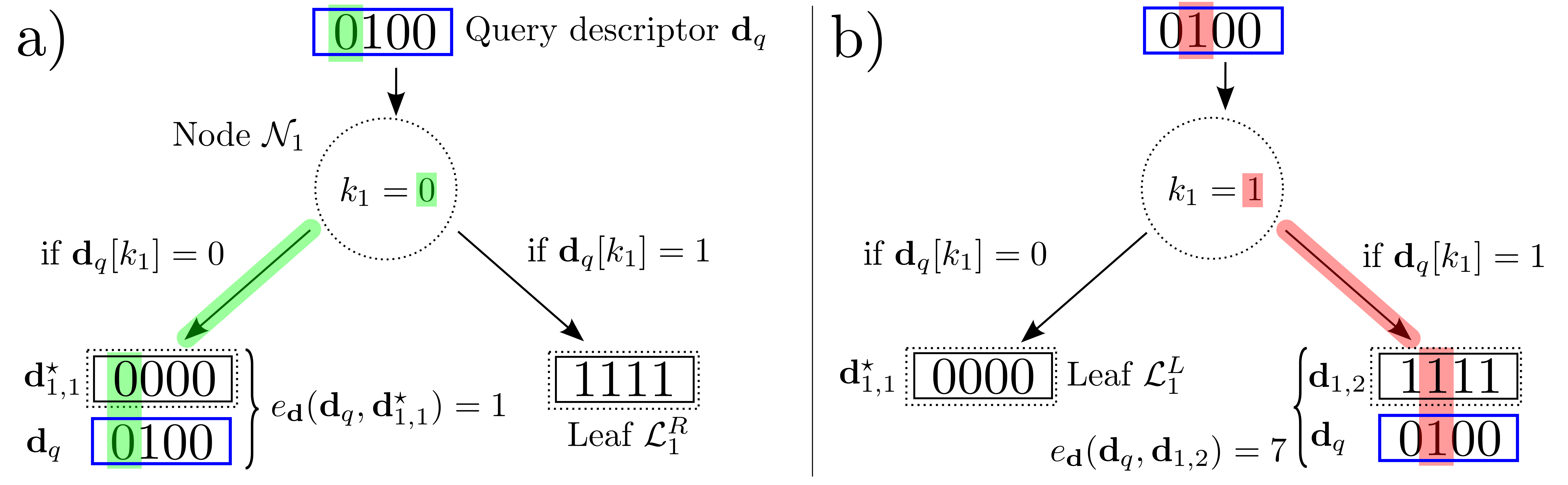}
  \vspace{-10pt}
  \caption{Search scenarios a) and b) for a small tree of depth $h=1$.
  The only configuration change between the scenarios is the value of $k_1$.
  In this example only a single descriptor is contained in each leaf.
  For $\querydescriptor$ the best matching descriptor is $\descriptor^\star_{1,1}$,
  which is found in a) but not in b).}
  \label{fig:hbst-matching}
  \vspace{-5pt}
\end{figure}

The most similar descriptor $\descriptor^\star_{i,j}$ to a query $\querydescriptor$ is stored in a leaf $\leaf^\star_i$.
This leaf is reached by traversing the tree, starting from the root. At each intermediate node $\node_i$  the search branches according to
$\querydescriptor[\bitindex]$.
Eventually, the search will end up in a leaf $\leaf^\star_i$. At this point
all leaf descriptors $\leafdescriptors$ stored in $\leaf^\star_i$ are sequentially scanned (BF matching) to seek for the best match
according to~\eqref{eq:descriptor-matching}.
\figref{fig:hbst-matching} illustrates two examples of the proposed search strategy.

Organizing $N_j$ descriptors in a balanced tree of depth $h$, results
in having an average of $\frac{N_j}{2^h}$ descriptors in a leaf.
Consequently, the time complexity of a search is:
\begin{equation}
  \complexity(h+\frac{N_j}{2^h})
  \label{eq:search-complexity}
\end{equation}
since $h$ operations are needed to find the leaf
and the descriptor matching in the leaf can be performed in $\frac{N_j}{2^h}$ steps.

If a query descriptor $\querydescriptor$ is already contained in the tree,
the search is guaranteed to correctly return the stored descriptor $\descriptor^\star_{i,j}$.
This, however does not hold for nearest neighbor searches when one is interested in
finding the descriptor in the tree that is similar to $\querydescriptor$.  
This is a consequence of the binary search procedure that preforms a greedy search based on
the bit index $\bitindex$ at each node.  Once a leaf is reached, only descriptors in that leaf are considered
as potential results. Thus we can say that in general the nearest neighbor search in the tree is
not ensured to be \emph{correct}.
In practice, however,  one is usually interested in
finding a descriptor $\descriptor_{i,j}$ that is \emph{similar enough} to $\querydescriptor$.
Hence incorrect matches are tolerated as long as they are not too far off w.r.t. $\querydescriptor$,
according to the metric $\descriptor_{i,j}:\distancedescriptor(\querydescriptor, \descriptor_{i,j}) < \tau$. 

If we want to retrieve \emph{all} descriptors that lay
within a certain distance
$\left\{\descriptor^\star_{i,j}:\distancedescriptor(\querydescriptor,
\descriptor_{i,j}) < \tau\right\}$, the search in the tree might be
not \emph{complete}. Incompleteness occurs when only a subset of the
feasible matches are returned from a search.  If a search is complete,
it is also correct.

\subsection{Completeness Analysis}
\label{sec:completeness-analysis}
A bounded nearest neighbor search for a query descriptor
$\querydescriptor$ and a threshold $\tau$ is said to be
\emph{complete} if all possible matching descriptors
$\{\descriptor_j^{(\tau,q)} \}$ such that
$\distancedescriptor(\querydescriptor, \descriptor_j^{(\tau,q)}) <
\tau$ are returned.  Given $\querydescriptor$, our search procedure
returns all descriptors $\{\descriptor_{i,j}^{(\tau,q)} \}$ in the
leaf $\leaf_i$ whose distance is below $\tau$.  These matching
descriptors are necessarily a subset of all feasible ones
$\{\descriptor_{i,j}^{(\tau,q)}\} \subset \{\descriptor_j^{(\tau,q)}
\}$. A straightforward measure of completeness for a \emph{single}
descriptor search is:
\begin{equation}
  \completeness_\tau(\querydescriptor)=\frac{|\{\descriptor_{i,j}^{(\tau,q)}\}|}{|\{\descriptor_j^{(\tau,q)}\}|} \in [0,1].
  \label{eq:completeness}
\end{equation}

Given a set of input descriptors $\inputdescriptors$, a set of query
descriptors $\querydescriptors$, a search threshold $\tau$ and a
search tree constructed from $\inputdescriptors$, we can evaluate
the \emph{mean completeness} $\overline{\completeness}_\tau(\querydescriptors)$ over all searches.
This gives us a meaningful measure of the overall completeness of our search.

Since the structure of the search tree is governed by the choice of
bit indices $\bitindices$, we conducted an experiment to evaluate how
the choice of $\bitindex$ influences the completeness, under different
thresholds $\tau$. Therefore we evaluated the resulting \emph{bitwise} mean completeness
$\overline{\completeness}_\tau(\bitindex)$ obtained by constructing
trees for every possible bit index $\bitindex$. Without loss of
generality we restricted our evaluation to a number of $n=\dimdescriptor$ trees $\{\tree^{(n)}\}$.
Each tree $\tree^{(n)}$ consists of only a root node
$\node^{(n)}_{1}$ with bit index $k^{(n)}_{1}=n$ and two single leafs
that partition the descriptors (similar to the tree described in~\figref{fig:hbst-matching}).

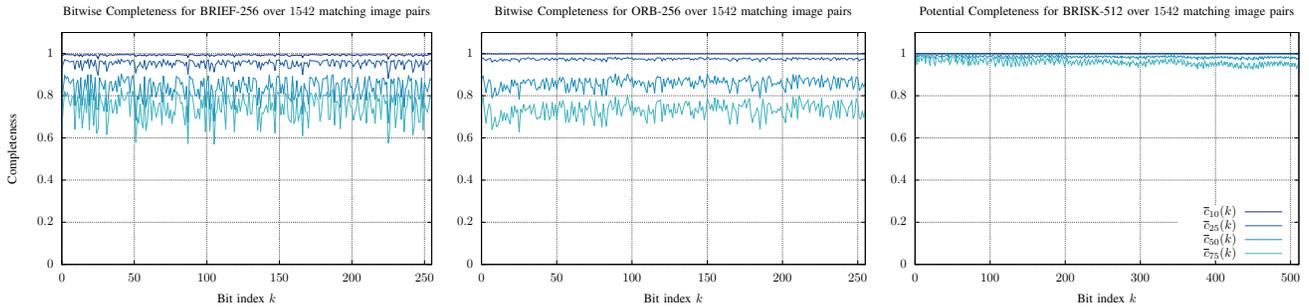
\begin{figure*}
\centering
  \begin{subfigure}{0.34\textwidth}
    \resizebox{\columnwidth}{!}{\begin{picture}(7200.00,5040.00)%
    \gdef\gplbacktext{}
    \gdef\gplfronttext{}
    \gplgaddtomacro\gplbacktext{%
      \csname LTb\endcsname%
      \put(814,704){\makebox(0,0)[r]{\strut{}$0$}}%
      \csname LTb\endcsname%
      \put(814,1372){\makebox(0,0)[r]{\strut{}$0.2$}}%
      \csname LTb\endcsname%
      \put(814,2040){\makebox(0,0)[r]{\strut{}$0.4$}}%
      \csname LTb\endcsname%
      \put(814,2709){\makebox(0,0)[r]{\strut{}$0.6$}}%
      \csname LTb\endcsname%
      \put(814,3377){\makebox(0,0)[r]{\strut{}$0.8$}}%
      \csname LTb\endcsname%
      \put(814,4045){\makebox(0,0)[r]{\strut{}$1$}}%
      \csname LTb\endcsname%
      \put(946,484){\makebox(0,0){\strut{}$0$}}%
      \csname LTb\endcsname%
      \put(2094,484){\makebox(0,0){\strut{}$50$}}%
      \csname LTb\endcsname%
      \put(3243,484){\makebox(0,0){\strut{}$100$}}%
      \csname LTb\endcsname%
      \put(4391,484){\makebox(0,0){\strut{}$150$}}%
      \csname LTb\endcsname%
      \put(5540,484){\makebox(0,0){\strut{}$200$}}%
      \csname LTb\endcsname%
      \put(6688,484){\makebox(0,0){\strut{}$250$}}%
    }%
    \gplgaddtomacro\gplfronttext{%
      \csname LTb\endcsname%
      \put(176,2541){\rotatebox{-270}{\makebox(0,0){\strut{}Completeness}}}%
      \put(3874,154){\makebox(0,0){\strut{}Bit index $k$}}%
      \put(3874,4709){\makebox(0,0){\strut{}Bitwise Completeness for BRIEF-256 over $1542$ matching image pairs}}%
    }%
    \gplbacktext
    \put(0,0){\includegraphics{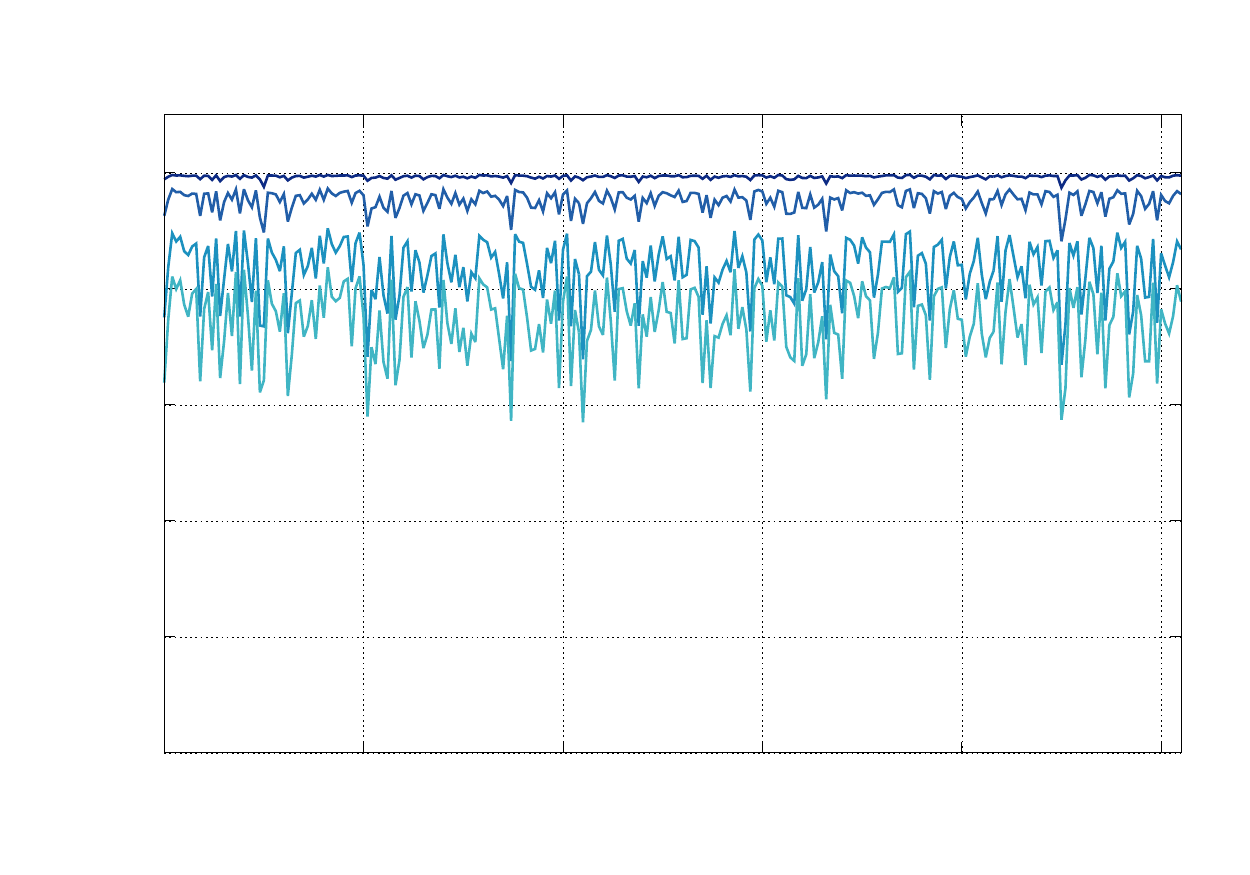}}%
    \gplfronttext
  \end{picture}}
  \end{subfigure}
  \hspace{-15pt}
  \begin{subfigure}{0.34\textwidth}
    \resizebox{\columnwidth}{!}{\begin{picture}(7200.00,5040.00)%
    \gdef\gplbacktext{}
    \gdef\gplfronttext{}
    \gplgaddtomacro\gplbacktext{%
      \csname LTb\endcsname%
      \put(594,704){\makebox(0,0)[r]{\strut{}$0$}}%
      \csname LTb\endcsname%
      \put(594,1372){\makebox(0,0)[r]{\strut{}$0.2$}}%
      \csname LTb\endcsname%
      \put(594,2040){\makebox(0,0)[r]{\strut{}$0.4$}}%
      \csname LTb\endcsname%
      \put(594,2709){\makebox(0,0)[r]{\strut{}$0.6$}}%
      \csname LTb\endcsname%
      \put(594,3377){\makebox(0,0)[r]{\strut{}$0.8$}}%
      \csname LTb\endcsname%
      \put(594,4045){\makebox(0,0)[r]{\strut{}$1$}}%
      \csname LTb\endcsname%
      \put(726,484){\makebox(0,0){\strut{}$0$}}%
      \csname LTb\endcsname%
      \put(1918,484){\makebox(0,0){\strut{}$50$}}%
      \csname LTb\endcsname%
      \put(3109,484){\makebox(0,0){\strut{}$100$}}%
      \csname LTb\endcsname%
      \put(4301,484){\makebox(0,0){\strut{}$150$}}%
      \csname LTb\endcsname%
      \put(5492,484){\makebox(0,0){\strut{}$200$}}%
      \csname LTb\endcsname%
      \put(6684,484){\makebox(0,0){\strut{}$250$}}%
    }%
    \gplgaddtomacro\gplfronttext{%
      \csname LTb\endcsname%
      \put(3764,154){\makebox(0,0){\strut{}Bit index $k$}}%
      \put(3764,4709){\makebox(0,0){\strut{}Bitwise Completeness for ORB-256 over $1542$ matching image pairs}}%
    }%
    \gplbacktext
    \put(0,0){\includegraphics{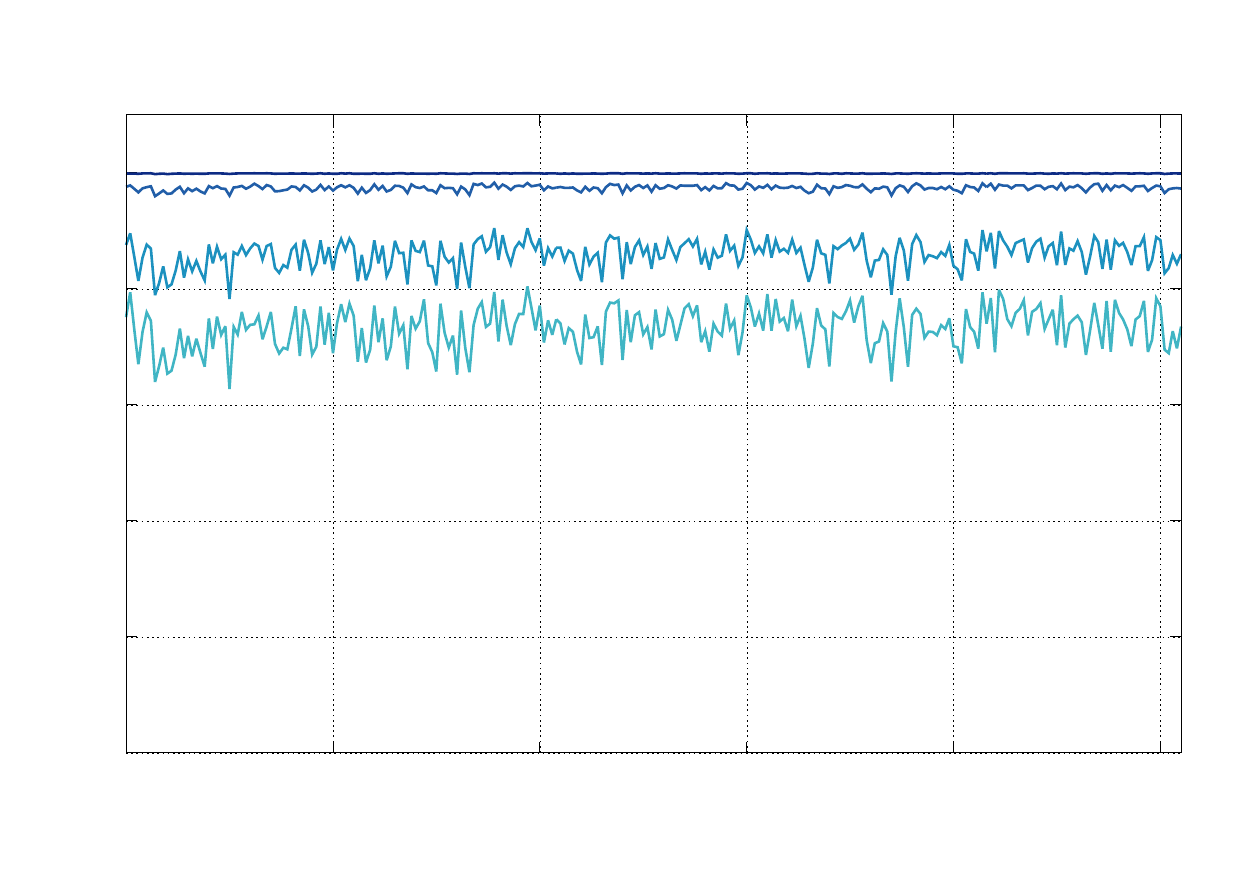}}%
    \gplfronttext
  \end{picture}}
  \end{subfigure}
  \hspace{-15pt}
  \begin{subfigure}{0.34\textwidth}
    \resizebox{\columnwidth}{!}{\begin{picture}(7200.00,5040.00)%
    \gdef\gplbacktext{}
    \gdef\gplfronttext{}
    \gplgaddtomacro\gplbacktext{%
      \csname LTb\endcsname%
      \put(594,704){\makebox(0,0)[r]{\strut{}$0$}}%
      \csname LTb\endcsname%
      \put(594,1372){\makebox(0,0)[r]{\strut{}$0.2$}}%
      \csname LTb\endcsname%
      \put(594,2040){\makebox(0,0)[r]{\strut{}$0.4$}}%
      \csname LTb\endcsname%
      \put(594,2709){\makebox(0,0)[r]{\strut{}$0.6$}}%
      \csname LTb\endcsname%
      \put(594,3377){\makebox(0,0)[r]{\strut{}$0.8$}}%
      \csname LTb\endcsname%
      \put(594,4045){\makebox(0,0)[r]{\strut{}$1$}}%
      \csname LTb\endcsname%
      \put(726,484){\makebox(0,0){\strut{}$0$}}%
      \csname LTb\endcsname%
      \put(1915,484){\makebox(0,0){\strut{}$100$}}%
      \csname LTb\endcsname%
      \put(3104,484){\makebox(0,0){\strut{}$200$}}%
      \csname LTb\endcsname%
      \put(4294,484){\makebox(0,0){\strut{}$300$}}%
      \csname LTb\endcsname%
      \put(5483,484){\makebox(0,0){\strut{}$400$}}%
      \csname LTb\endcsname%
      \put(6672,484){\makebox(0,0){\strut{}$500$}}%
    }%
    \gplgaddtomacro\gplfronttext{%
      \csname LTb\endcsname%
      \put(3764,154){\makebox(0,0){\strut{}Bit index $k$}}%
      \put(3764,4709){\makebox(0,0){\strut{}Potential Completeness for BRISK-512 over $1542$ matching image pairs}}%
      \csname LTb\endcsname%
      \put(5816,1537){\makebox(0,0)[r]{\strut{}$\overline{c}_{10}(k)$}}%
      \csname LTb\endcsname%
      \put(5816,1317){\makebox(0,0)[r]{\strut{}$\overline{c}_{25}(k)$}}%
      \csname LTb\endcsname%
      \put(5816,1097){\makebox(0,0)[r]{\strut{}$\overline{c}_{50}(k)$}}%
      \csname LTb\endcsname%
      \put(5816,877){\makebox(0,0)[r]{\strut{}$\overline{c}_{75}(k)$}}%
    }%
    \gplbacktext
    \put(0,0){\includegraphics{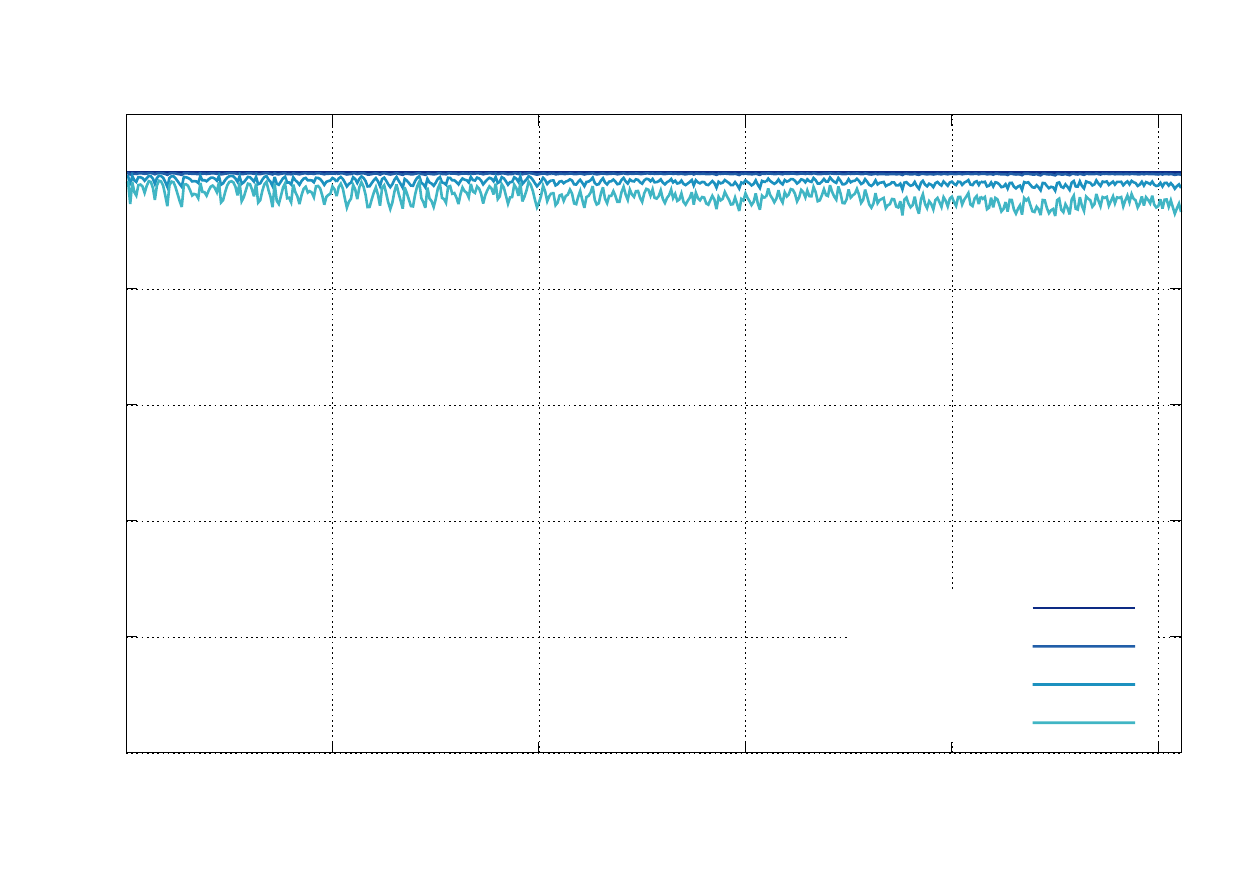}}%
    \gplfronttext
  \end{picture}}
  \end{subfigure}
  
  \caption{Bitwise mean completeness $\overline{\completeness}_\tau(\bitindex)$ for matching thresholds $\tau\in\left\{10, 25, 50, 75\right\}$ and 3 binary descriptor types.
  A number of $N_\descriptor=1000$ descriptors has been extracted per image.
  The ground truth for this experiment consisted of 1542 image pairs, corresponding to over 1.5 million descriptors in the database.
  The colorization and legend based on $\tau$ is identical for all plots and can be inspected in the rightmost figure.
  Note that the BRISK-512 descriptor has $\descriptorsize=512$ and therefore the considered matching threshold $\tau$ is much more restrictive with respect to BRIEF-256 and ORB-256}
  \label{fig:completeness-split}
  \vspace{-15pt}
\end{figure*}

In~\figref{fig:completeness-split} we report the results of our
bitwise completeness analysis on sequence 00 of the KITTI benchmark
dataset~\cite{geiger2012we}.  A broader analysis, featuring also FREAK and A-KAZE
descriptors as well as many other datasets, is available on the project website.

Based on the results shown in~\figref{fig:completeness-split} two facts are evident:
\begin{itemize}
\item The choice of the bit index $\bitindex$ does not substantially influence the
      mean completeness $\overline{\completeness}_\tau(\bitindex)$ of the nearest neighbor search. This behavior is similar
      for different types of binary descriptors.
\item The greater the threshold $\tau$, the lower the mean completeness $\overline{\completeness}_\tau(\bitindex)$ becomes.
\end{itemize}

\begin{figure}[ht!]
  \centering
  \vspace{-10pt}
  \resizebox{\columnwidth}{!}{\begin{picture}(7200.00,5040.00)%
    \gdef\gplbacktext{}
    \gdef\gplfronttext{}
    \gplgaddtomacro\gplbacktext{%
      \csname LTb\endcsname%
      \put(814,704){\makebox(0,0)[r]{\strut{}$0$}}%
      \csname LTb\endcsname%
      \put(814,1439){\makebox(0,0)[r]{\strut{}$0.2$}}%
      \csname LTb\endcsname%
      \put(814,2174){\makebox(0,0)[r]{\strut{}$0.4$}}%
      \csname LTb\endcsname%
      \put(814,2909){\makebox(0,0)[r]{\strut{}$0.6$}}%
      \csname LTb\endcsname%
      \put(814,3644){\makebox(0,0)[r]{\strut{}$0.8$}}%
      \csname LTb\endcsname%
      \put(814,4379){\makebox(0,0)[r]{\strut{}$1$}}%
      \csname LTb\endcsname%
      \put(946,484){\makebox(0,0){\strut{}$0$}}%
      \csname LTb\endcsname%
      \put(1503,484){\makebox(0,0){\strut{}$2$}}%
      \csname LTb\endcsname%
      \put(2059,484){\makebox(0,0){\strut{}$4$}}%
      \csname LTb\endcsname%
      \put(2616,484){\makebox(0,0){\strut{}$6$}}%
      \csname LTb\endcsname%
      \put(3173,484){\makebox(0,0){\strut{}$8$}}%
      \csname LTb\endcsname%
      \put(3729,484){\makebox(0,0){\strut{}$10$}}%
      \csname LTb\endcsname%
      \put(4286,484){\makebox(0,0){\strut{}$12$}}%
      \csname LTb\endcsname%
      \put(4842,484){\makebox(0,0){\strut{}$14$}}%
      \csname LTb\endcsname%
      \put(5399,484){\makebox(0,0){\strut{}$16$}}%
    }%
    \gplgaddtomacro\gplfronttext{%
      \csname LTb\endcsname%
      \put(176,2541){\rotatebox{-270}{\makebox(0,0){\strut{}Completeness}}}%
      \put(3172,154){\makebox(0,0){\strut{}Tree depth $h$}}%
      \put(3172,4709){\makebox(0,0){\strut{}Mean Completeness for different depths (BRIEF-256)}}%
      \csname LTb\endcsname%
      \put(6212,4269){\makebox(0,0)[r]{\strut{}$\overline c_{10}(h)$}}%
      \csname LTb\endcsname%
      \put(6212,4049){\makebox(0,0)[r]{\strut{}$\hat{c}_{10}(h)$}}%
      \csname LTb\endcsname%
      \put(6212,3829){\makebox(0,0)[r]{\strut{}$\overline c_{25}(h)$}}%
      \csname LTb\endcsname%
      \put(6212,3609){\makebox(0,0)[r]{\strut{}$\hat{c}_{25}(h)$}}%
      \csname LTb\endcsname%
      \put(6212,3389){\makebox(0,0)[r]{\strut{}$\overline c_{50}(h)$}}%
      \csname LTb\endcsname%
      \put(6212,3169){\makebox(0,0)[r]{\strut{}$\hat{c}_{50}(h)$}}%
      \csname LTb\endcsname%
      \put(6212,2949){\makebox(0,0)[r]{\strut{}$\overline c_{75}(h)$}}%
      \csname LTb\endcsname%
      \put(6212,2729){\makebox(0,0)[r]{\strut{}$\hat{c}_{75}(h)$}}%
    }%
    \gplbacktext
    \put(0,0){\includegraphics{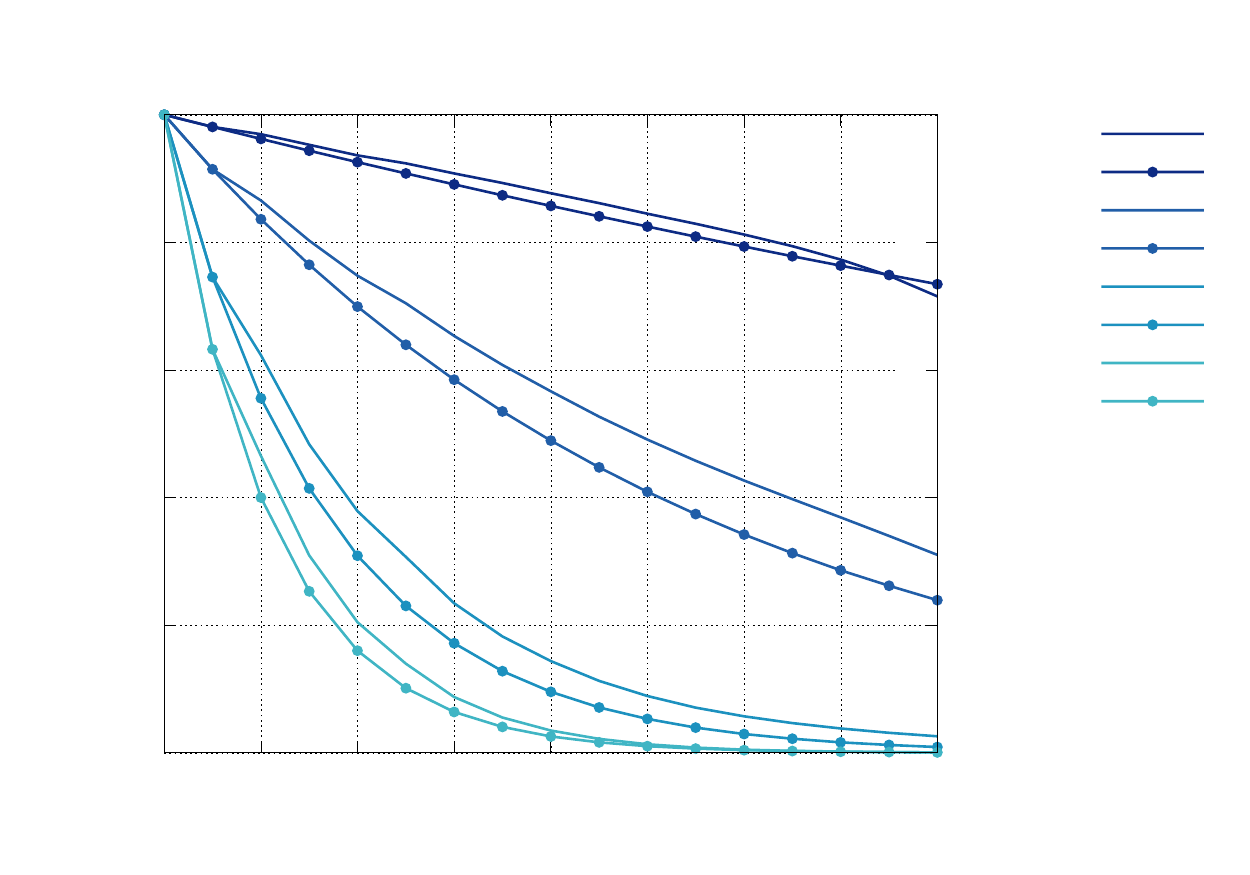}}%
    \gplfronttext
  \end{picture}}
  \vspace{-15pt}
  \caption{Measured and predicted mean completeness:
  $\overline\completeness_\tau(h)$ and $\hat{\completeness}_\tau(h)$
  for increasing depth $h=\left\{0, 1, .., 16\right\}$ at varying matching thresholds $\tau$.
  This experiment was conducted for 1542 matching images of the KITTI sequence 00 using BRIEF-256 descriptors.
  The values for $\overline\completeness_\tau(1)$ correspond
  to the mean values of $\overline\completeness_\tau(k)$ reported in \figref{fig:completeness-split}.}
  \label{fig:completeness-evolution}
  \vspace{-0pt}
\end{figure}
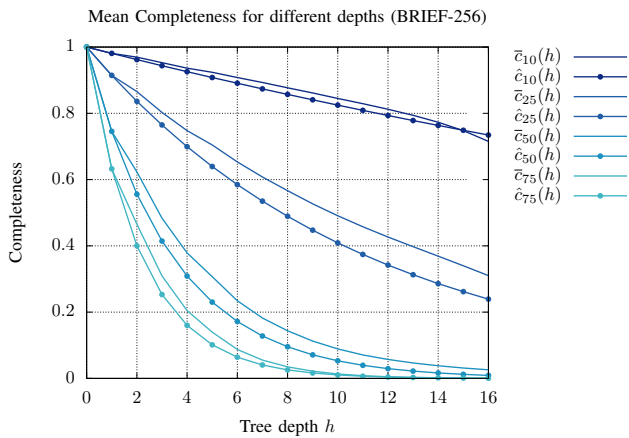

Whereas the above experiment (\figref{fig:completeness-split}) only considers trees of depth $h=1$,
its results can be used to predict the evolution of the completeness
as the depth of the tree increases further.
Let $\overline\completeness_\tau$ be the average completeness over all bit indices $\bitindices$ at threshold $\tau$,
for a tree having depth $1$.
Performing a search on a tree at depth $h > 1$, would result in applying the decision rule $\descriptor[\bitindex]$
exactly $h$ times, and each decision would result in a potential loss of completeness according to~\eqref{eq:completeness}.
Assuming that $\overline\completeness_\tau$ is evaluated on a
representative set of query and input descriptors,
we expect that $\overline\completeness_\tau$
does not change significantly on other tree \emph{levels} as well.
A tree level is a single view of a node $\node_i$ and two leafs
which can be inspected at any depth in the tree by collapsing the left and right subtree of $\node_i$.
Thus we predict the completeness at depth $h$ as:
\begin{equation}
  \hat\completeness_\tau(h)={\overline\completeness_\tau}^h\in\left[0, 1\right].
  \label{eq:completeness-prediction}
\end{equation}
To confirm our conjuncture, we conducted a second experiment,
where we organize the input descriptor set in a sequence of balanced trees $\left\{\tree^{(h)}\right\}$ with increasing depths $h=\left\{0, 1, .., 16\right\}$.
We then repeated the evaluation conducted in the previous experiment (\figref{fig:completeness-split}),
by computing the average completeness of all queries for all depths.
\figref{fig:completeness-evolution} reports the results of this evaluation.

From this analysis we further conclude that:
\begin{itemize}
\item The completeness $\overline\completeness_\tau(h)$ decreases exponentially with increasing depth $h$ of the search tree.
\item \eqref{eq:completeness-prediction} captures the relation between depth of the tree and completeness reasonably well.
\end{itemize}

Summarizing the results of both experiments,
we found that the bit index $\bitindex$ does not significantly affect the mean completeness $\overline\completeness_\tau$.
Increasing the tree depth $h$ on the other hand drastically reduces $\overline\completeness_\tau$.

\subsection{Balanced Tree Construction}
\label{sec:balanced-tree-construction}
In this section we describe how to organize a set of descriptors in a
\emph{balanced} tree of depth $h$.
Considering~\eqref{eq:search-complexity} and~\eqref{eq:completeness-prediction},
for a given threshold $\tau$, we have a trade-off between search time and completeness.
Higher values of $h$ will result in increased search speed at the cost of a reduced completeness.
These results however, hold only in the case of balanced trees, and both
search speed and completeness will decrease when the tree becomes unbalanced.

A straightforward strategy to build a balanced tree from a set of input
descriptors $\inputdescriptors$ consists in recursively splitting the
current input descriptors evenly.
Since the structure of the tree is governed by the choice of $\bitindices$, to
achieve an even partitioning of $\inputdescriptors$, we choose
the bit index for which the bit $\inputdescriptor[\bitindex]$ is ``on'' for half of $\inputdescriptors$ for \emph{every} node $\node_i$.
The chosen bit index $k_i^\star$ will therefore be the one whose \emph{mean value} among all descriptors
is the \emph{closest} to 0.5:
\begin{equation}
  k_i^\star = \argmin_\bitindex {\Big\vert} 0.5- \frac{1}{N_j} \sum_j \inputdescriptor[\bitindex] {\Big\vert}.
  \label{eq:decision-rule}
\end{equation}
Note that when selecting $\bitindex$ we have to neglect
all the indices that have been used in the nodes ancestors.
If the minimized norm in~\eqref{eq:decision-rule} is below a certain threshold $\delta_{max}$,
we say that the mean value is \emph{close enough} to 0.5 and pick $k_i^\star$ for splitting.
In case that no such mean value is available,
we do not split the descriptors and the recursion stops.

Constructing a tree of depth $h$ for $N_j$ descriptors according to~\eqref{eq:decision-rule} has a
complexity of $\complexity(N_j \cdot h)$. In typical applications
such as SLAM, $N_j$ grows significantly for every new image, as new
descriptors are added to the set. Therefore constructing the tree
from scratch for all descriptors of all images for every new image
quickly leads to runtimes not adequate for real-time applications.
To overcome this computational limitation we propose an alternative
strategy to \emph{insert} new images (i.e. descriptors) into an
\emph{existing} tree.

\subsection{Incremental Tree Construction}
\label{sec:approach-insertion}
\begin{figure}[ht!]
  \centering
  \vspace{-0pt}
  \includegraphics[width=\columnwidth]{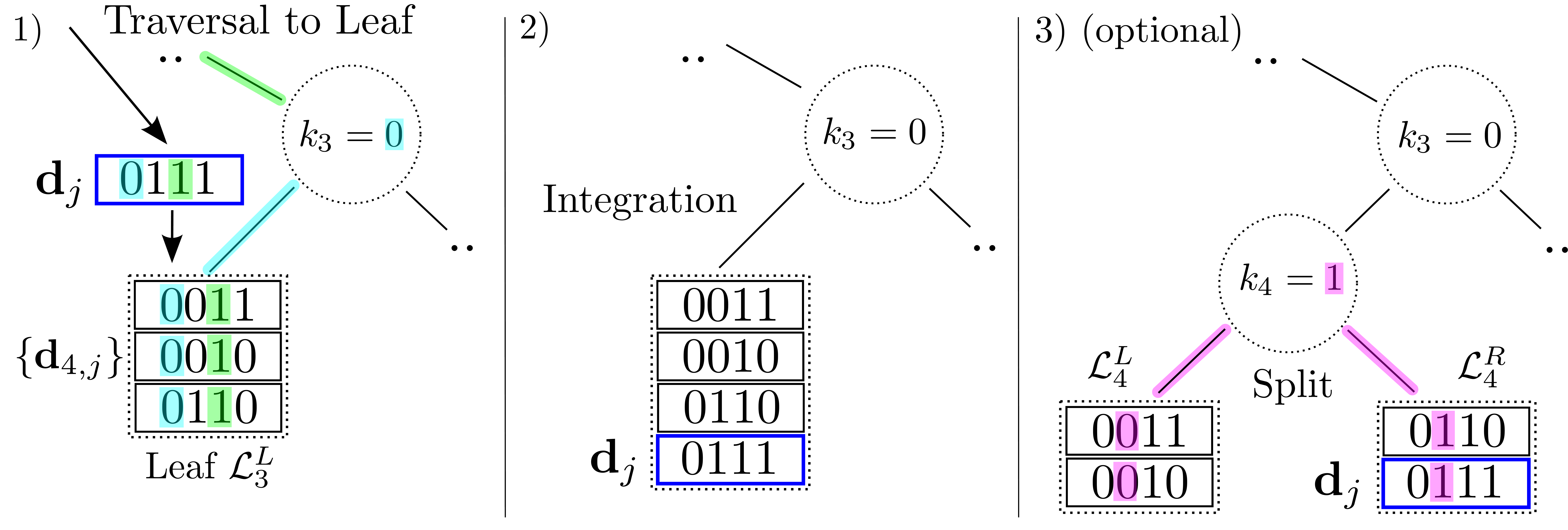}
  \vspace{-15pt}
  \caption{Descriptor insertion procedure for a single input descriptor $\inputdescriptor$.
  Only the affected part of the tree is shown.
  Step 1) The leaf $\leaf_i$ containing the most similar descriptor(s) to $\inputdescriptor$ is found.
  Step 2) $\inputdescriptor$ is integrated into the leaf descriptor set $\left\{\descriptor_{4,j}\right\}$.
  Step 3) If the leaf becomes ``too big'': $N_{4,j}>\maxleafsize$, it breaks into two child leafs and becomes an intermediate node.
  In this example we set the maximum leaf size to $\maxleafsize=3$.}
  \label{fig:hbst-insertion}
  \vspace{-5pt}
\end{figure}

In this section we describe an alternative strategy that allows to
augment an initial tree with additional descriptors while limiting its depth.
The idea is to accumulate descriptors in
a leaf until a number $\maxleafsize$ (maximum leaf size) is reached.
Whereas hierarchical clustering trees~\cite{2012-muja-fast-matching}
use the maximum leaf size as termination criterion for the clustering process,
we on the other hand evaluate it to determine if a clustering (i.e. splitting) is necessary.
When the maximum leaf size is exceeded, 
we say that the leaf $\leaf_i$ becomes ``too large'',
and we turn the leaf in an intermediate node $\node_i$. 
The bit index $\bitindex$ for $\node_i$ is
selected according to the criterion in~\eqref{eq:decision-rule}, and
the descriptors previously contained in $\leaf_i$ are organized in two
new leafs spawning from $\node_i$.
\figref{fig:hbst-insertion} illustrates the proposed procedure.

\begin{figure}[ht!]
  \centering
  \vspace{-70pt}
  \resizebox{\columnwidth}{!}{\begin{picture}(7200.00,5040.00)%
    \gdef\gplbacktext{}
    \gdef\gplfronttext{}
    \gplgaddtomacro\gplbacktext{%
      \csname LTb\endcsname%
      \put(682,704){\makebox(0,0)[r]{\strut{}$0$}}%
      \csname LTb\endcsname%
      \put(682,1131){\makebox(0,0)[r]{\strut{}$5$}}%
      \csname LTb\endcsname%
      \put(682,1558){\makebox(0,0)[r]{\strut{}$10$}}%
      \csname LTb\endcsname%
      \put(682,1985){\makebox(0,0)[r]{\strut{}$15$}}%
      \csname LTb\endcsname%
      \put(682,2412){\makebox(0,0)[r]{\strut{}$20$}}%
      \csname LTb\endcsname%
      \put(682,2839){\makebox(0,0)[r]{\strut{}$25$}}%
      \csname LTb\endcsname%
      \put(814,484){\makebox(0,0){\strut{}$0$}}%
      \csname LTb\endcsname%
      \put(1716,484){\makebox(0,0){\strut{}$5000$}}%
      \csname LTb\endcsname%
      \put(2618,484){\makebox(0,0){\strut{}$10000$}}%
      \csname LTb\endcsname%
      \put(3520,484){\makebox(0,0){\strut{}$15000$}}%
      \csname LTb\endcsname%
      \put(4422,484){\makebox(0,0){\strut{}$20000$}}%
      \csname LTb\endcsname%
      \put(5324,484){\makebox(0,0){\strut{}$25000$}}%
      \csname LTb\endcsname%
      \put(6226,484){\makebox(0,0){\strut{}$30000$}}%
    }%
    \gplgaddtomacro\gplfronttext{%
      \csname LTb\endcsname%
      \put(176,1771){\rotatebox{-270}{\makebox(0,0){\strut{}Tree depth $h$}}}%
      \put(3808,154){\makebox(0,0){\strut{}Number of inserted images}}%
      \put(3808,3169){\makebox(0,0){\strut{}Tree depths on St. Lucia (BRIEF-256)}}%
      \csname LTb\endcsname%
      \put(6080,1097){\makebox(0,0)[r]{\strut{}Standard deviation}}%
      \csname LTb\endcsname%
      \put(6080,877){\makebox(0,0)[r]{\strut{}Mean depth}}%
    }%
    \gplbacktext
    \put(0,0){\includegraphics{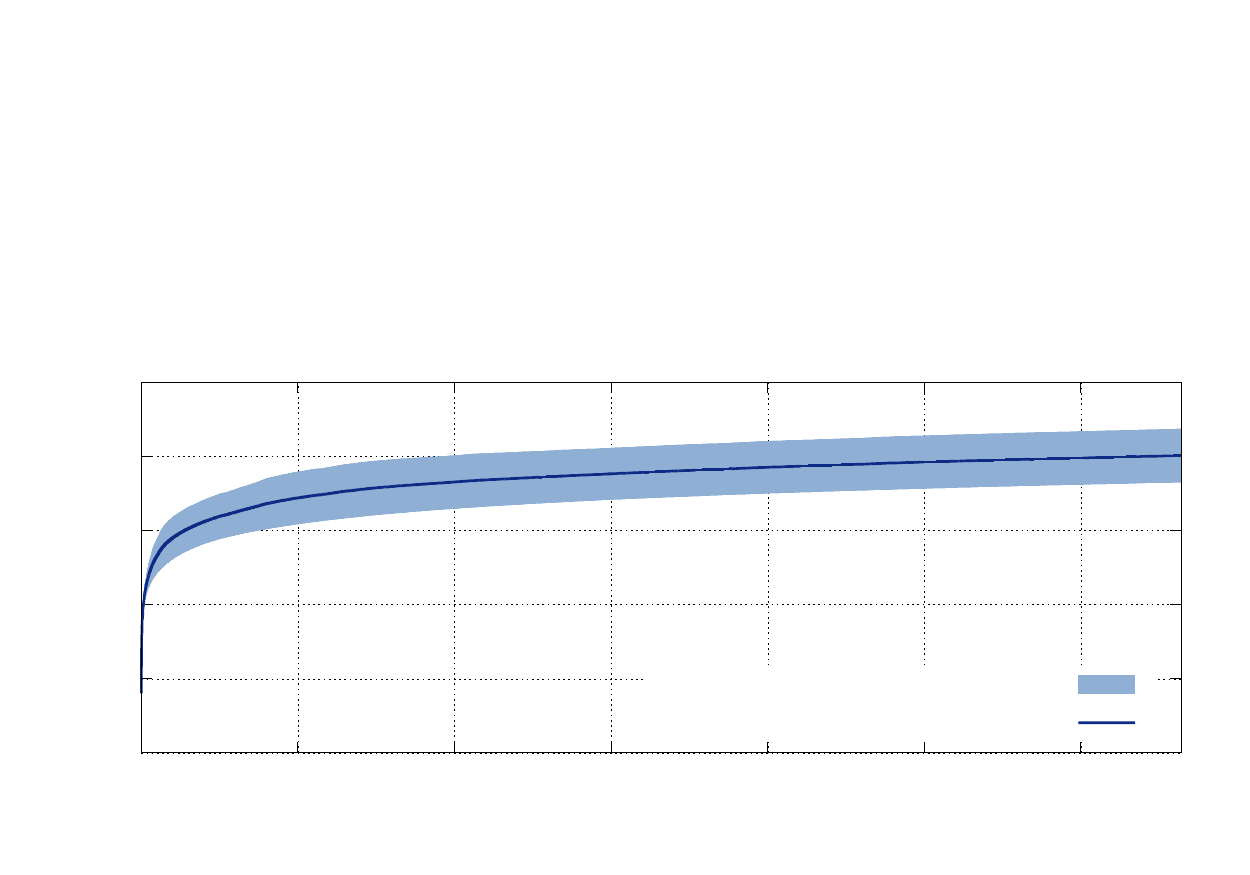}}%
    \gplfronttext
  \end{picture}}\\
  \vspace{-3pt}
  \caption{Mean and standard deviation of the tree depth $h$ for increasing numbers of sequentially inserted images.
  $N_\descriptor=1000$ BRIEF-256 descriptors were extracted for each of the 33'197 images in the dataset (resulting over 30 million inserted descriptors).
  For this experiment we set $\tau=25, \delta_{max}=0.1$ and $\maxleafsize=100$.}
  \label{fig:leaf-evolution}
  \vspace{-10pt}
\end{figure}
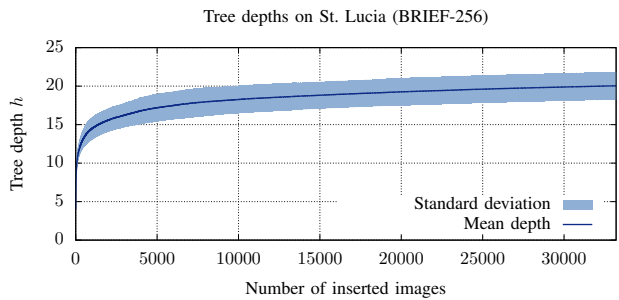

Notably the tree traversal needed to find $\leaf_i$ is the same as for the search.
This enables us to perform both search and insertion at the same time.
Albeit this straightforward insertion technique does not guarantee a balanced tree,
it succeeds in limiting the depth of the tree as shown in~\figref{fig:leaf-evolution}.

Note that using re-balancing structures such as Red-Black trees to organize the descriptors
is not straightforward in our case. Since the constraint that a bit index would appear
at most once in a path from root to leaf would be violated by moving nodes.
In our approach, no tree re-balancing is performed as we are able to enforce a desired balance to a satisfiable degree using
the parameter $\delta_{max}$ (\secref{sec:balanced-tree-construction}).


To enable image retrieval,
we augment each stored descriptor with the index of the image
from which it was extracted.
This allows us to implement a voting scheme for image retrieval at no additional cost.

\section{Experiments}
\label{sec:experiments}
In this section we report the results of a comparative evaluation of
our approach with several state-of-the-art methods (\secref{sec:compared-approaches}).
We measure the image retrieval accuracy and the runtime of each method on multiple publicly available
datasets (\secref{sec:datasets}).
To quantify the accuracy of image retrieval,
we extract a VPR ground truth on which images should match for the analyzed datasets,
using a brute-force offline procedure (\secref{sec:ground-truth-computation}). 
For space reasons and due to the high number of considered datasets,
we report the achieved accuracy using the maximum $F_1$ score,
which is a single number summarizing the well known Precision-Recall curves (\secref{sec:precision-recall}).

For each dataset and each approach we process the images sequentially.
Every time a new image is acquired, the approaches are queried for image retrieval.
Based on the reported image matches and on the provided ground truth we then can estimate
precision, recall and $F_1$ score. 
The database is subsequently augmented by inserting the new image,
so that it can be returned as a match in future queries.
We gather runtime information by measuring
the average time $\overline{t}$ spent for both of these operations for each image.

\subsection{Compared Approaches}
\label{sec:compared-approaches}
Our comparison has been conducted on the following state-of-the-art image retrieval approaches:
\begin{itemize}
\item BF: The classical brute-force approach. We utilize the current OpenCV3 implementation.
  BF is expected to achieve the highest precision and recall,
  while requiring the highest processing time $\overline{t}$ per image.

\item FLANN-LSH: We utilize the current OpenCV3 implementation of FLANN with LSH indexing.
  The LSH index is built using the parameters:
  $\mathrm{table\_number}=10$, $\mathrm{key\_size}=20$ and $\mathrm{multi\_probe\_level}=0$.

\item DBoW2-DI: We used the DBoW2 approach \emph{with} Direct Indexing.
  Image matches are pruned by decreasing number of matches obtained through
  descriptor matching using the provided DBoW2 indices.
  DBoW2 was run with parameters: $\mathrm{use\_di}=true$ and $\mathrm{di\_levels}=2$.

\item DBoW2-SO: DBoW2 \emph{without} direct indexing.
  Accordingly the parameters are: $\mathrm{use\_di}=false$.
  This configuration does not report matching descriptors but only matching images (based on image Score Only).

\item HBST-10: HBST is the approach proposed in this paper, with parameters:
  $\delta_{max}=0.1$, $\maxleafsize=10$. 

\item HBST-50: Same as above but with an extended maximum leaf size of $\maxleafsize=50$. 
  HBST-50 is designed to provide a higher accuracy than HBST-10
  at the price of a higher processing time $\overline{t}$.
\end{itemize}
For all approaches we considered a maximum descriptor matching distance of $\tau=25$
and we extracted for each image $N_\descriptor=1000$ BRIEF-256 descriptors.
All results were obtained on the same machine,
running Ubuntu 16.04.3 with an Intel i7-7700K CPU@4.2GHz and 32GB of RAM@4.1GHz.
A more extensive evaluation featuring various binary descriptor types
(e.g. ORB, BRISK, FREAK and A-KAZE) is available on the project website.

\subsection{Datasets}
\label{sec:datasets}
\begin{figure}[ht!]
  \centering
  \vspace{-10pt}
  \begin{subfigure}{0.49\columnwidth}
    \includegraphics[width=\columnwidth]{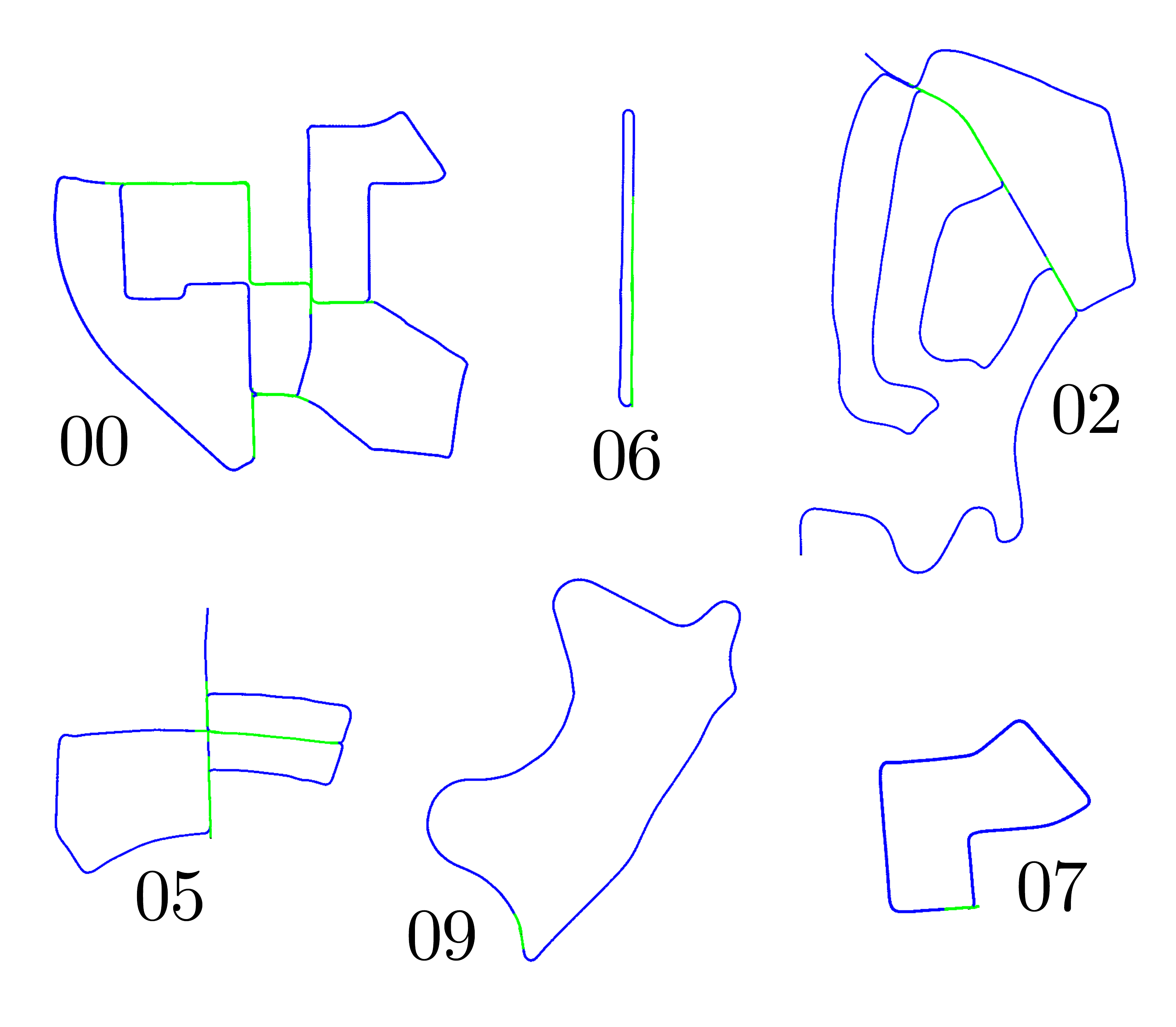}
    \caption{KITTI: 14.6 km, 15'756 images.}
    \vspace{-0pt}
  \end{subfigure}
  \begin{subfigure}{0.49\columnwidth}
    \includegraphics[width=\columnwidth]{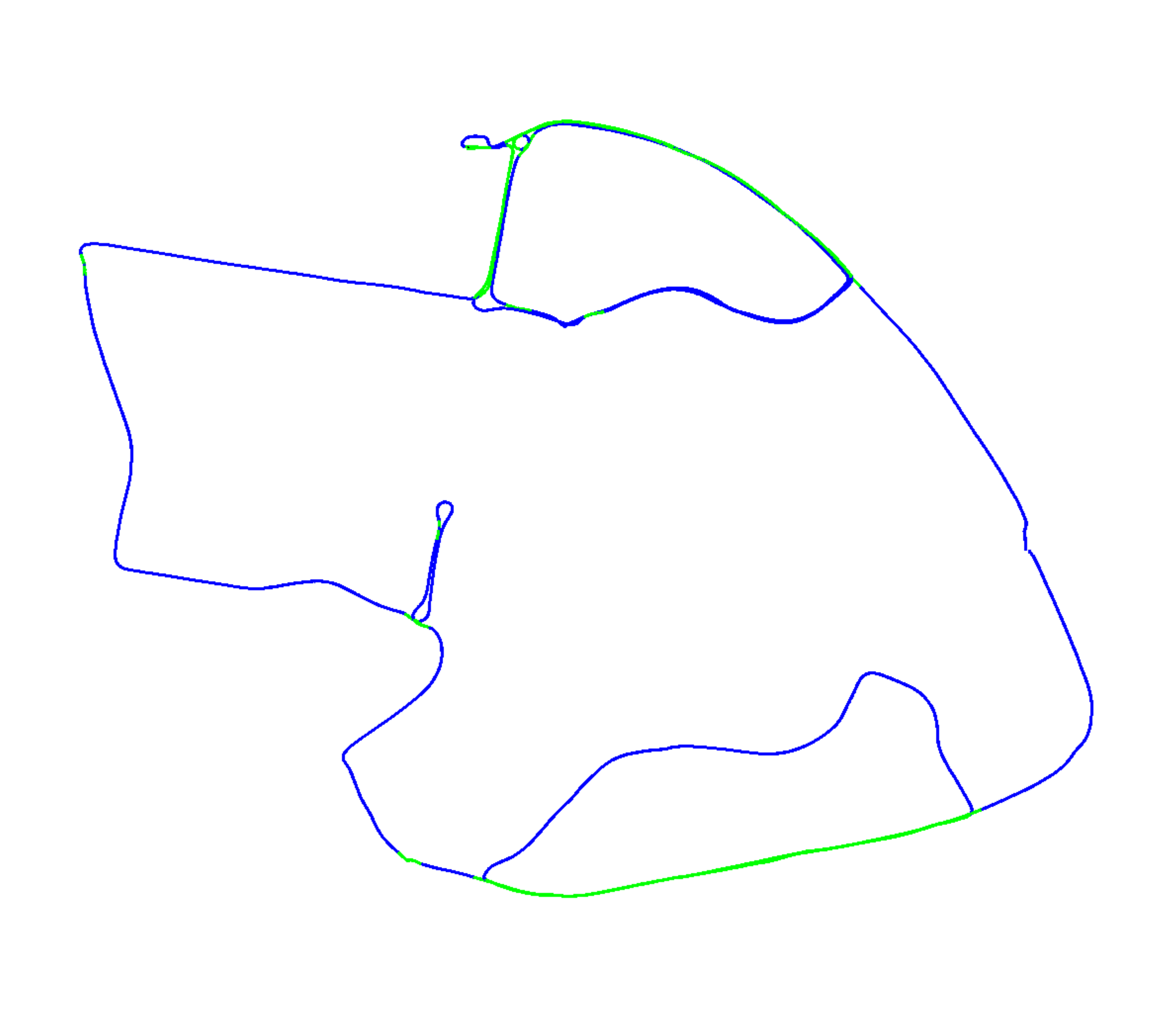}
    \caption{St. Lucia: 9.5 km, 33'197 images.}
    \vspace{-0pt}
  \end{subfigure}
  \caption{Selected datasets sequences with acquisition trajectories in blue.
  Respective matching image segments (defined by the VPR ground truth of \secref{sec:ground-truth-computation}) are highlighted in green.}
  \label{fig:datasets}
  \vspace{-10pt}
\end{figure}

We performed our result evaluation on 4 publicly available large-scale visual SLAM datasets:
KITTI~\cite{geiger2012we}, M\'alaga~\cite{2013-blanco-malaga}, St. Lucia~\cite{warren2010} and Oxford~\cite{2017-oxford-dataset}.
Each dataset contains multiple sequences with thousands of images.
In \figref{fig:datasets} we show an aerial view of the robot trajectories in these sequences.
For space reasons, we report in this paper only the results of KITTI and St. Lucia,
being in line with the other datasets.
The results of M\'alaga and Oxford can be inspected on the project website.

\subsection{Ground Truth Computation}
\label{sec:ground-truth-computation}
Obtaining the ground truth for image retrieval is a crucial aspect of our evaluation.
To this extent we employ a brute-force approach aided by the ground truth camera pose information available in the datasets.
We report a \emph{match} (true positive) between a query $\image_q$ and an image $\image_i$ in the database,
whenever \emph{all} of the following criteria are met:
\begin{enumerate}
\item The fields of view at which the images were acquired must overlap,
  and the camera positions have to be close.
  This occurs when two images are acquired at positions closer than 10 meters,
  and the optical axes of the cameras have an angular distance below 20 degrees.
  
\item Since all approaches are designed to approximate the BF accuracy,
  we require that matching images are supported by a minimum number of matching descriptors.
  This test is passed when more than 10\% of the descriptors are within the matching threshold $\tau=25$.
  
\item To confirm the usability of returned descriptor matches for image registration,
  we perform a geometric validation for the keypoint correspondences $\left<\keypoint_q, \keypoint_j\right>$.
  A correspondence $\left<\keypoint_q, \keypoint_j\right>$ is valid,
  if the essential constraint $\keypoint_q^\top\mathbf{E}\keypoint_j=0$ is approached~\cite{zisserman}.

\end{enumerate}
The tool we used to generate such a ground truth for image matches is available
online\footnote{\scriptsize Benchmark project:
  \url{www.gitlab.com/srrg-software/srrg_bench}}.
The subset of matches that passes our criteria forms the set of \emph{ground truth matches}.

\subsection{Precision, Recall and the F1 score}
\label{sec:precision-recall}
To determine the reliability of a place recognition approach
one generally measures the resulting \emph{Precision} and \emph{Recall} statistics.
The first statistic being:
\begin{equation}
Precision = \frac{\#~correctly~reported~associations}{\#~total~reported~associations} \in \left[0,1\right]. \notag
\end{equation}
Here $\#~correctly~reported~associations$ is the subset of matches reported that are also in the ground truth set,
while the $\#~total~reported~associations$ are all matches returned.
To evaluate the completeness we also consider:
\begin{equation}
Recall = \frac{\#~correctly~reported~associations}{\#~total~possible~associations} \in \left[0,1\right]. \notag
\end{equation}
Here $\#~total~possible~associations$ are all associations in the ground truth set.
The $F_1$ score is a compact measure that combines Precision and Recall in a single value: 
\begin{equation}
F_1 = 2\cdot\frac{Precision \cdot Recall}{Precision+Recall} \in \left[0,1\right] \notag
\end{equation}
The \emph{maximum} $F_1$ score obtained by a method represents the \emph{best} tradeoff between Precision and Recall.
The higher the $F_1$ score, the more accurate \emph{and} complete is an approach.

\subsection{Results}
\label{sec:results}
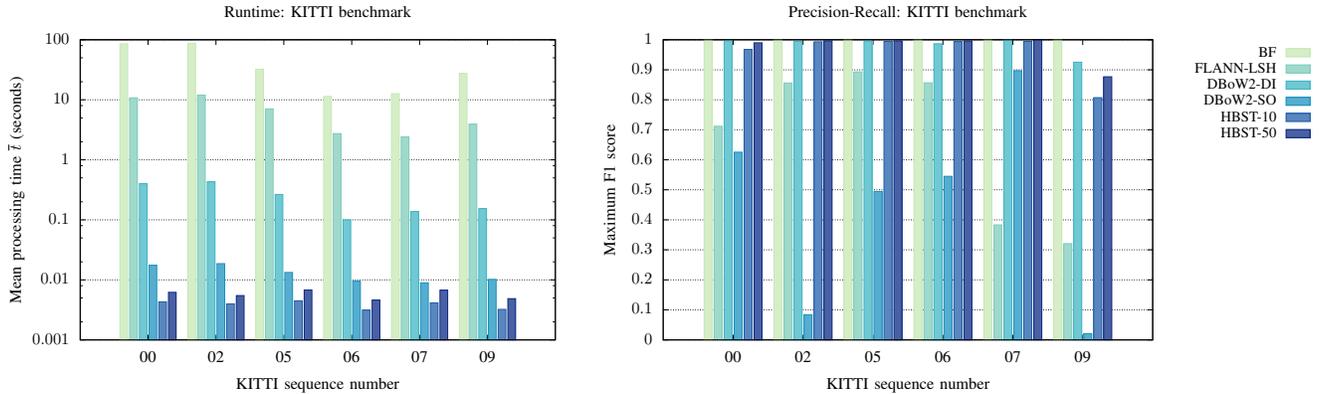
\begin{figure*}
  \centering
  \vspace{-0pt}
  \begin{subfigure}[]{0.44\textwidth}
    \resizebox{\columnwidth}{!}{\begin{picture}(7200.00,5040.00)%
    \gdef\gplbacktext{}
    \gdef\gplfronttext{}
    \gplgaddtomacro\gplbacktext{%
      \csname LTb\endcsname%
      \put(858,704){\makebox(0,0)[r]{\strut{}$0.001$}}%
      \csname LTb\endcsname%
      \put(858,1439){\makebox(0,0)[r]{\strut{}$0.01$}}%
      \csname LTb\endcsname%
      \put(858,2174){\makebox(0,0)[r]{\strut{}$0.1$}}%
      \csname LTb\endcsname%
      \put(858,2909){\makebox(0,0)[r]{\strut{}$1$}}%
      \csname LTb\endcsname%
      \put(858,3644){\makebox(0,0)[r]{\strut{}$10$}}%
      \csname LTb\endcsname%
      \put(858,4379){\makebox(0,0)[r]{\strut{}$100$}}%
      \put(1820,484){\makebox(0,0){\strut{}00}}%
      \put(2651,484){\makebox(0,0){\strut{}02}}%
      \put(3481,484){\makebox(0,0){\strut{}05}}%
      \put(4312,484){\makebox(0,0){\strut{}06}}%
      \put(5142,484){\makebox(0,0){\strut{}07}}%
      \put(5973,484){\makebox(0,0){\strut{}09}}%
    }%
    \gplgaddtomacro\gplfronttext{%
      \csname LTb\endcsname%
      \put(220,2541){\rotatebox{-270}{\makebox(0,0){\strut{}Mean processing time $\overline{t}$ (seconds)}}}%
      \put(3896,154){\makebox(0,0){\strut{}KITTI sequence number}}%
      \put(3896,4709){\makebox(0,0){\strut{}Runtime: KITTI benchmark}}%
    }%
    \gplbacktext
    \put(0,0){\includegraphics{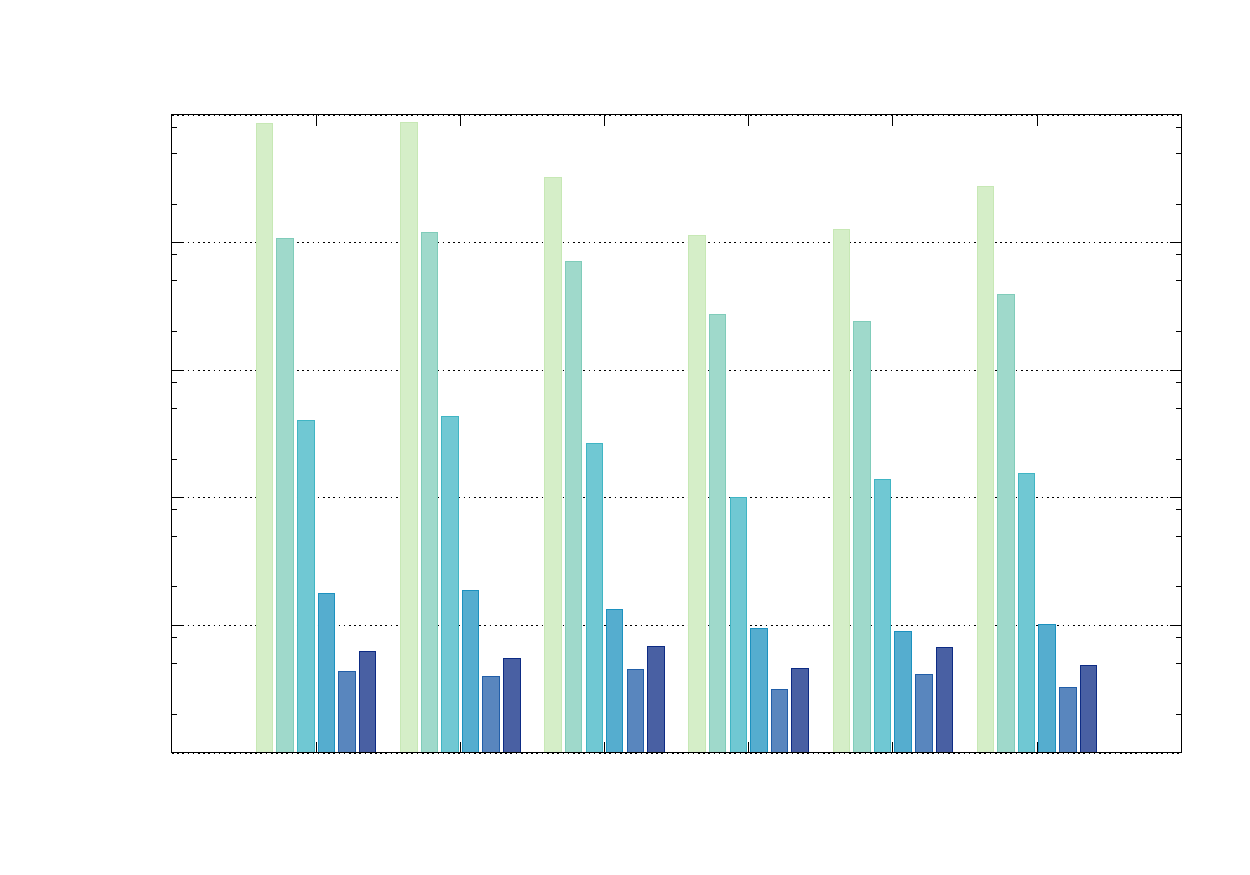}}%
    \gplfronttext
  \end{picture}}
  \end{subfigure}
  \begin{subfigure}[]{0.44\textwidth}
    \resizebox{\columnwidth}{!}{\begin{picture}(7200.00,5040.00)%
    \gdef\gplbacktext{}
    \gdef\gplfronttext{}
    \gplgaddtomacro\gplbacktext{%
      \csname LTb\endcsname%
      \put(682,704){\makebox(0,0)[r]{\strut{}$0$}}%
      \csname LTb\endcsname%
      \put(682,1072){\makebox(0,0)[r]{\strut{}$0.1$}}%
      \csname LTb\endcsname%
      \put(682,1439){\makebox(0,0)[r]{\strut{}$0.2$}}%
      \csname LTb\endcsname%
      \put(682,1807){\makebox(0,0)[r]{\strut{}$0.3$}}%
      \csname LTb\endcsname%
      \put(682,2174){\makebox(0,0)[r]{\strut{}$0.4$}}%
      \csname LTb\endcsname%
      \put(682,2542){\makebox(0,0)[r]{\strut{}$0.5$}}%
      \csname LTb\endcsname%
      \put(682,2909){\makebox(0,0)[r]{\strut{}$0.6$}}%
      \csname LTb\endcsname%
      \put(682,3277){\makebox(0,0)[r]{\strut{}$0.7$}}%
      \csname LTb\endcsname%
      \put(682,3644){\makebox(0,0)[r]{\strut{}$0.8$}}%
      \csname LTb\endcsname%
      \put(682,4011){\makebox(0,0)[r]{\strut{}$0.9$}}%
      \csname LTb\endcsname%
      \put(682,4379){\makebox(0,0)[r]{\strut{}$1$}}%
      \put(1670,484){\makebox(0,0){\strut{}00}}%
      \put(2525,484){\makebox(0,0){\strut{}02}}%
      \put(3381,484){\makebox(0,0){\strut{}05}}%
      \put(4236,484){\makebox(0,0){\strut{}06}}%
      \put(5092,484){\makebox(0,0){\strut{}07}}%
      \put(5947,484){\makebox(0,0){\strut{}09}}%
    }%
    \gplgaddtomacro\gplfronttext{%
      \csname LTb\endcsname%
      \put(176,2541){\rotatebox{-270}{\makebox(0,0){\strut{}Maximum F1 score}}}%
      \put(3808,154){\makebox(0,0){\strut{}KITTI sequence number}}%
      \put(3808,4709){\makebox(0,0){\strut{}Precision-Recall: KITTI benchmark}}%
    }%
    \gplbacktext
    \put(0,0){\includegraphics{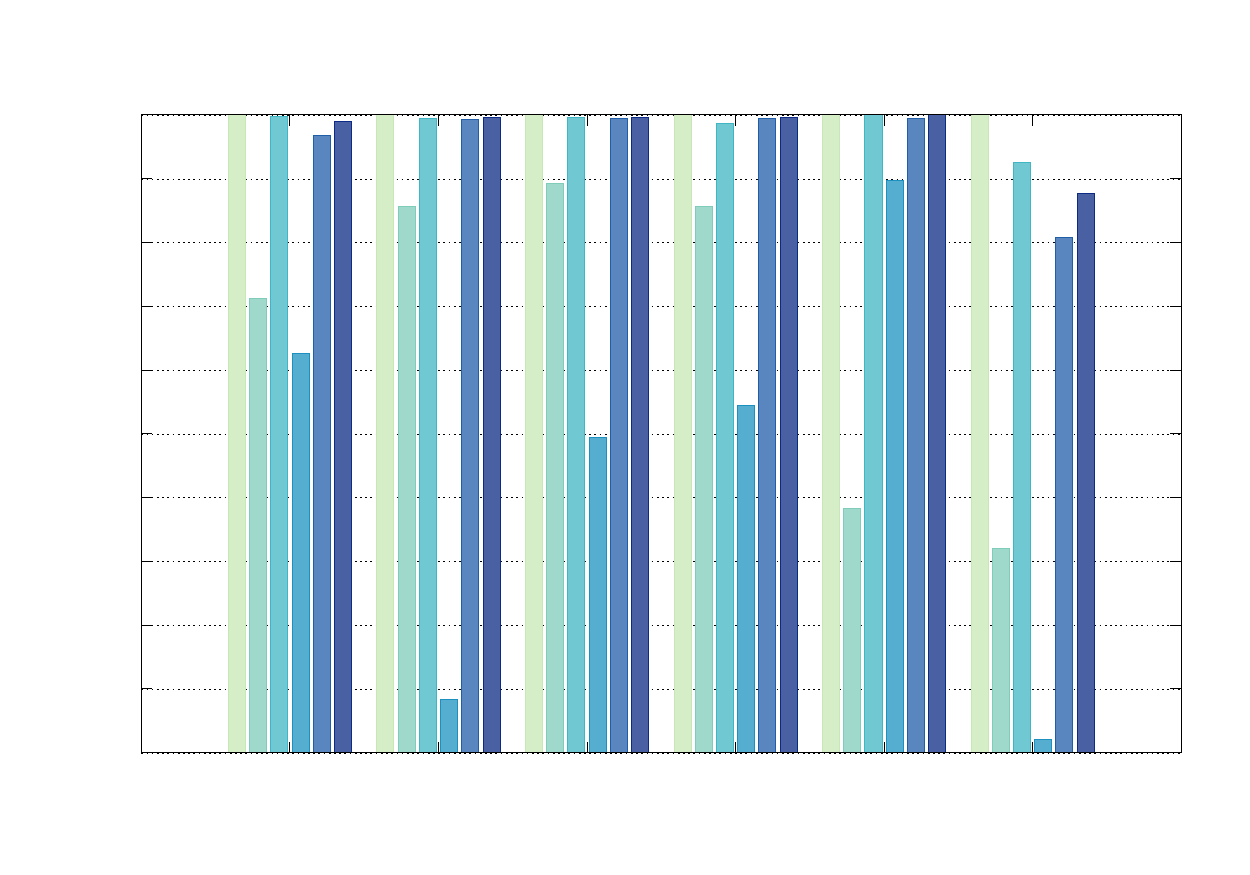}}%
    \gplfronttext
  \end{picture}}
  \end{subfigure}
  \begin{subfigure}{0.1\textwidth}
    \adjustbox{width=\columnwidth, trim=10pt 0pt 260pt 40pt, clip}{\begin{picture}(7200.00,5040.00)%
    \gdef\gplbacktext{}
    \gdef\gplfronttext{}
    \gplgaddtomacro\gplbacktext{%
    }%
    \gplgaddtomacro\gplfronttext{%
      \csname LTb\endcsname%
      \put(1320,4011){\makebox(0,0)[r]{\strut{}BF}}%
      \csname LTb\endcsname%
      \put(1320,3791){\makebox(0,0)[r]{\strut{}FLANN-LSH}}%
      \csname LTb\endcsname%
      \put(1320,3571){\makebox(0,0)[r]{\strut{}DBoW2-DI}}%
      \csname LTb\endcsname%
      \put(1320,3351){\makebox(0,0)[r]{\strut{}DBoW2-SO}}%
      \csname LTb\endcsname%
      \put(1320,3131){\makebox(0,0)[r]{\strut{}HBST-10}}%
      \csname LTb\endcsname%
      \put(1320,2911){\makebox(0,0)[r]{\strut{}HBST-50}}%
    }%
    \gplbacktext
    \put(0,0){\includegraphics{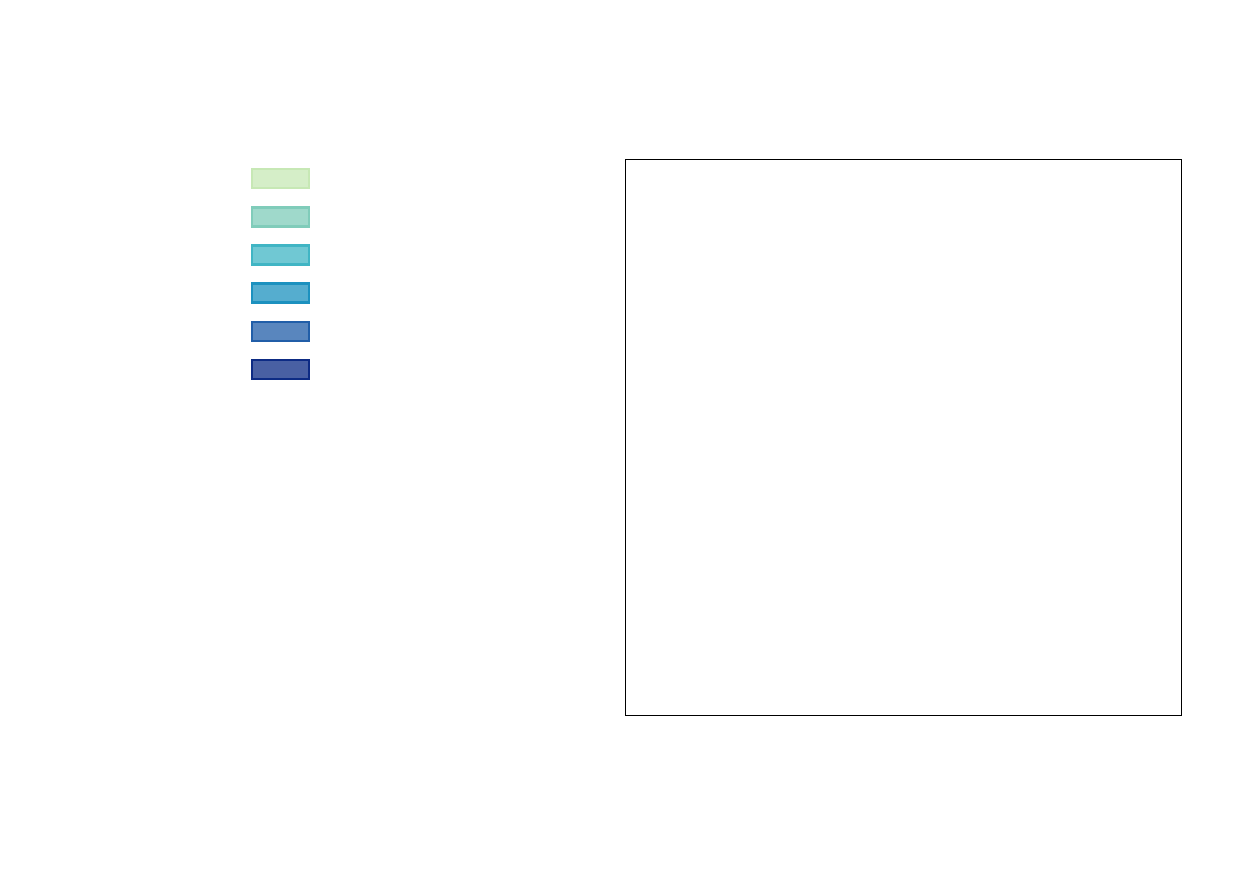}}%
    \gplfronttext
  \end{picture}}
  \end{subfigure}
  \vspace{-5pt}
  \caption{KITTI Visual Odometry/SLAM Evaluation 2012: Large-scale urban and rural environments in Germany.
  We acquired a number of $N_\descriptor=1000$ BRIEF-256 descriptors per image in every sequence. 
  All approaches were evaluated on a single core. Note that the runtime axis is logarithmic.}
  \label{fig:benchmark-kitti}
  \vspace{-10pt}
\end{figure*}

In \figref{fig:benchmark-kitti} we report the results of all
approaches on KITTI. We observed the following for each compared approach:
\begin{itemize}
  \item BF: Not surprisingly, BF is clearly the most accurate, at the cost
    of a higher computation that grows linearly with the number of inserted
    images. BF prohibits real-time execution after 10 to 20 images.
  \item FLANN-LSH: It generally achieves decent $F_1$ scores between DBoW2-SO and HBST-10.
    Its high computational requirements are not adequate for a real-time application in our scenario.
  \item DBOW2-DI: The BoW approach achieves the best $F_1$ score after BF, at a
    computational cost that grows mildly with the number of images inserted.
    Yet it is two orders of magnitude slower than HBST.
  \item DBOW2-SO: The pure histogram comparison (Score Only) used with this settings
    leads to the poorest $F_1$ score.
    However it is the fastest approach after HBST.
  \item HBST-10: Our approach achieves accuracy between FLANN-LSH and DBOW2-DI,
    while it is by far the fastest approach compared.
  \item HBST-50: As expected, HBST-50 achieves a higher accuracy than HBST-10
    while being slightly slower. 
\end{itemize}

\begin{figure}[ht!]
  \hspace{2pt}
  \vspace{-0pt}
  \resizebox{\columnwidth}{!}{\begin{picture}(7200.00,5040.00)%
    \gdef\gplbacktext{}%
    \gdef\gplfronttext{}%
    \gplgaddtomacro\gplbacktext{%
      \csname LTb\endcsname%
      \put(588,704){\makebox(0,0)[r]{\strut{}$0.001$}}%
      \csname LTb\endcsname%
      \put(588,1317){\makebox(0,0)[r]{\strut{}$0.01$}}%
      \csname LTb\endcsname%
      \put(588,1929){\makebox(0,0)[r]{\strut{}$0.1$}}%
      \csname LTb\endcsname%
      \put(588,2542){\makebox(0,0)[r]{\strut{}$1$}}%
      \csname LTb\endcsname%
      \put(588,3154){\makebox(0,0)[r]{\strut{}$10$}}%
      \csname LTb\endcsname%
      \put(588,3767){\makebox(0,0)[r]{\strut{}$100$}}%
      \csname LTb\endcsname%
      \put(588,4379){\makebox(0,0)[r]{\strut{}$1000$}}%
      \csname LTb\endcsname%
      \put(720,484){\makebox(0,0){\strut{}$0$}}%
      \csname LTb\endcsname%
      \put(1418,484){\makebox(0,0){\strut{}$10000$}}%
      \csname LTb\endcsname%
      \put(2116,484){\makebox(0,0){\strut{}$20000$}}%
      \csname LTb\endcsname%
      \put(2814,484){\makebox(0,0){\strut{}$30000$}}%
    }%
    \gplgaddtomacro\gplfronttext{%
      \csname LTb\endcsname%
      \put(-50,2541){\rotatebox{-270}{\makebox(0,0){\strut{}Processing time $t_i$ (seconds)}}}%
      \put(1871,154){\makebox(0,0){\strut{}Image number $i$}}%
      \put(1871,4709){\makebox(0,0){\strut{}Runtime: St. Lucia}}%
    }%
    \gplgaddtomacro\gplbacktext{%
      \csname LTb\endcsname%
      \put(3240,484){\makebox(0,0){\strut{}$0$}}%
      \csname LTb\endcsname%
      \put(3888,484){\makebox(0,0){\strut{}$0.2$}}%
      \csname LTb\endcsname%
      \put(4536,484){\makebox(0,0){\strut{}$0.4$}}%
      \csname LTb\endcsname%
      \put(5183,484){\makebox(0,0){\strut{}$0.6$}}%
      \csname LTb\endcsname%
      \put(5831,484){\makebox(0,0){\strut{}$0.8$}}%
      \csname LTb\endcsname%
      \put(6479,484){\makebox(0,0){\strut{}$1$}}%
      \csname LTb\endcsname%
      \put(6611,704){\makebox(0,0)[l]{\strut{}$0$}}%
      \csname LTb\endcsname%
      \put(6611,1439){\makebox(0,0)[l]{\strut{}$0.2$}}%
      \csname LTb\endcsname%
      \put(6611,2174){\makebox(0,0)[l]{\strut{}$0.4$}}%
      \csname LTb\endcsname%
      \put(6611,2909){\makebox(0,0)[l]{\strut{}$0.6$}}%
      \csname LTb\endcsname%
      \put(6611,3644){\makebox(0,0)[l]{\strut{}$0.8$}}%
      \csname LTb\endcsname%
      \put(6611,4379){\makebox(0,0)[l]{\strut{}$1$}}%
    }%
    \gplgaddtomacro\gplfronttext{%
      \csname LTb\endcsname%
      \put(7116,2541){\rotatebox{-270}{\makebox(0,0){\strut{}Precision (correct/reported associations)}}}%
      \put(4859,154){\makebox(0,0){\strut{}Recall (correct/possible associations)}}%
      \put(4859,4709){\makebox(0,0){\strut{}Precision-Recall: St. Lucia}}%
      \csname LTb\endcsname%
      \put(3963,1977){\makebox(0,0)[l]{\strut{}\small BF}}%
      \csname LTb\endcsname%
      \put(3963,1757){\makebox(0,0)[l]{\strut{}\small FLANN-LSH}}%
      \csname LTb\endcsname%
      \put(3963,1537){\makebox(0,0)[l]{\strut{}\small DBoW2-DI}}%
      \csname LTb\endcsname%
      \put(3963,1317){\makebox(0,0)[l]{\strut{}\small DBoW2-SO}}%
      \csname LTb\endcsname%
      \put(3963,1097){\makebox(0,0)[l]{\strut{}\small HBST-10}}%
      \csname LTb\endcsname%
      \put(3963,877){\makebox(0,0)[l]{\strut{}\small HBST-50}}%
    }%
    \gplbacktext
    \put(0,0){\includegraphics{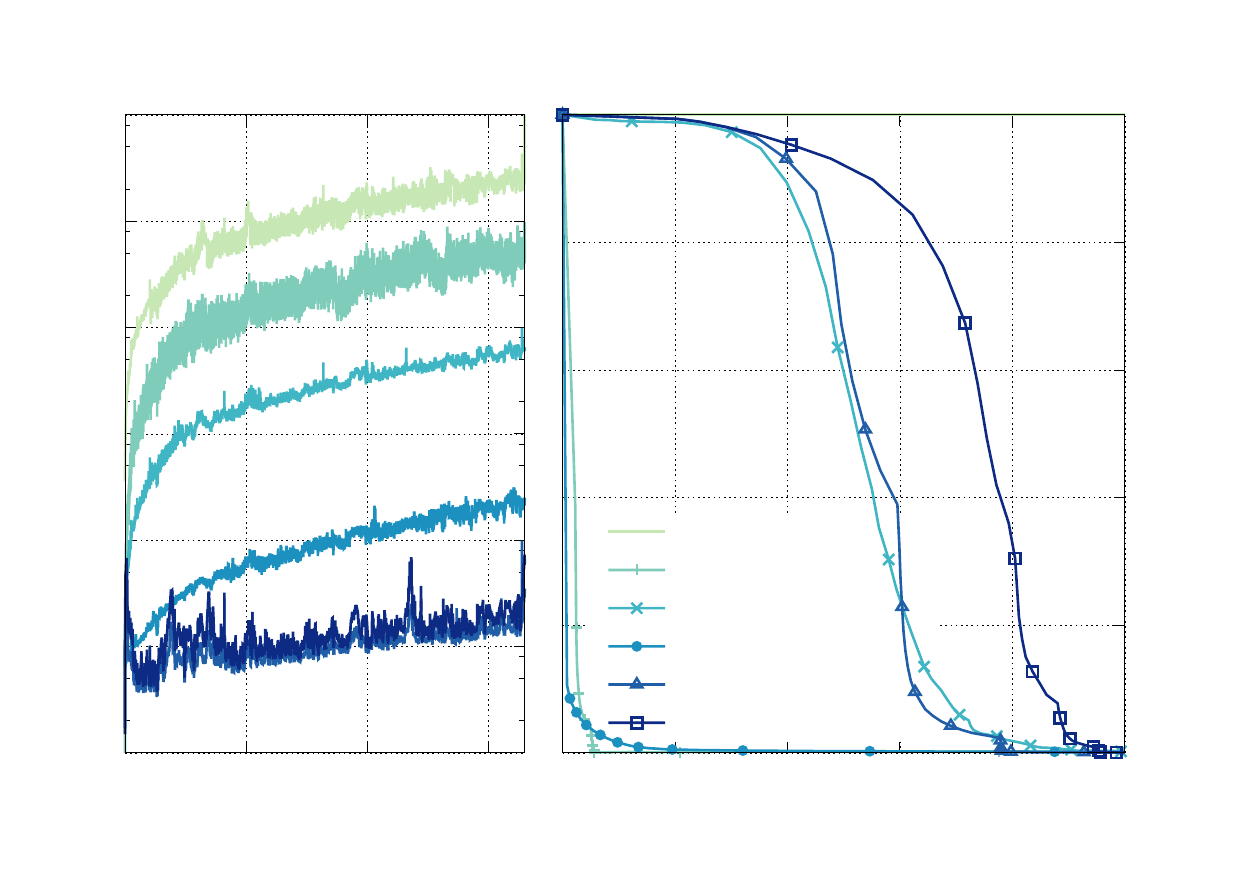}}%
    \gplfronttext
  \end{picture}}
  \vspace{-15pt}
  \caption{UQ St. Lucia Stereo Vehicular Dataset: Wide-ranging urban environment in Australia. With a total number of 33'197 images.
  $N_\descriptor=1000$ BRIEF-256 descriptors were computed for each image.}
  \label{fig:benchmark-lucia}
  \vspace{0pt}
\end{figure}
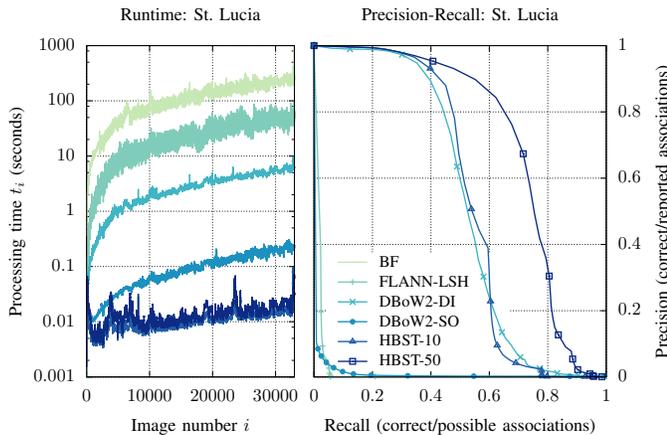

In \figref{fig:benchmark-lucia} we present a more detailed analysis
performed on a \emph{single} sequence with 33'197 images.
Here we show the Runtime and Precision-Recall curves of all approaches.
FLANN-LSH and DBoW2-SO fail due to the large, incrementally built database.
Both report many false positives, drastically reducing accuracy.
DBoW2-DI achieves acceptable accuracy, using descriptor matching to prune reported image matches.
Our method (HBST-10, HBST-50) outperforms all other approaches considered in this scenario.

\section{Conclusions}
\label{sec:conclusions}
In this paper we present a binary feature based search tree approach for Visual Place Recognition.
We conducted an analysis of the behavior of binary descriptors.
Based on this analysis we provide an approach that
can address descriptor matching and image retrieval.
While retaining an adequate accuracy,
our approach significantly outperforms state-of-the-art methods in terms of computational speed.
All of our results were obtained on publicly available datasets and
can be reproduced using the released open-source library.

\bibliographystyle{ieeetr}
\bibliography{robots}

\end{document}

%% file: notation.tex
\usepackage{amsopn}

\newcommand{\bp}{\mathbf{p}}

\DeclareMathOperator*{\argmin}{argmin}

\newcommand{\complexity}{\mathcal{O}}